\documentclass[review]{elsarticle}
\graphicspath{ {./figures/} }
\usepackage{hyperref}
\usepackage{float}
\usepackage{verbatim} %comments
\usepackage{apalike}
\restylefloat{figure}
\floatstyle{plaintop} %table caption at top
\restylefloat{table}
\usepackage{multirow,booktabs,color,soul,threeparttable}
\definecolor{hl}{rgb}{0.75,0.75,0.75}
\sethlcolor{hl}
\usepackage{graphicx}
\usepackage{subfigure}
\usepackage{amsmath}
\usepackage{stfloats}
\usepackage{multirow}
\usepackage{verbatim}
\usepackage{array}
\usepackage{marvosym}
\usepackage{soul}
\soulregister\cite7 % 针对\cite命令
\soulregister\citep7 % 针对\citep命令
\soulregister\citet7 % 针对\citet命令\soulregister\ref7 % 针对\ref命令
\soulregister\pageref7 % 针对\pageref命令
\usepackage[table,xcdraw]{xcolor}
\usepackage[ruled,vlined,linesnumbered]{algorithm2e} 
\usepackage{colortbl}

\journal{Expert Systems with Applications}

\biboptions{authoryear}

\begin{document}
\begin{frontmatter}

%The title page must contain the title of the paper and the full name/full affiliation with country/e-mail address for each author and co-author of the manuscript. Please make sure you have included all elements listed below with your manuscript submission.

\begin{titlepage}
\begin{center}
\vspace*{1cm}

\textbf{Ranking Constraints via Topological Dual-Directional Search in Evolutionary Multi-Objective Optimization}

\vspace{1.5cm}

% Author names and affiliations
Ruiqing Sun$^a$, Dawei Feng$^a$, Sheng Qi$^b$, Xing Zhou$^c$, Lianghao Li$^d$, Bo Ding$^a$, Yijie Wang$^{a,*}$, Rui Wang$^b$, Huaimin Wang$^a$

\vspace{0.5cm}

\begin{flushleft}
\small
$^a$ National Key Laboratory of Parallel and Distributed Computing, College of Computer Science and Technology, National University of Defense Technology, Changsha 410000, P.R. China \\
$^b$ College of System Engineering, National University of Defense Technology, Changsha 410000, P.R. China \\
$^c$ College of Intelligence Science and Technology, National University of Defense Technology, Changsha 410073, P.R. China \\
$^d$ State Key Laboratory of Complex \& Critical Software Environment, College of Information and Communication, National University of Defense Technology, Wuhan 430019, P.R. China

\vspace{0.8cm}
\textbf{Corresponding author:} \\
Yijie Wang \\
National Key Laboratory of Parallel and Distributed Computing, College of Computer Science and Technology, National University of Defense Technology, Changsha 410000, P.R. China \\
Email: wangyijie@nudt.edu.cn

\vspace{0.5cm}
\small
\end{flushleft}
\end{center}
\end{titlepage}

\title{Ranking Constraints via Topological Dual-Directional Search in Evolutionary Multi-Objective Optimization}

\author[label1]{Ruiqing Sun}
\ead{sunny0331@foxmail.com}

\author[label1]{Dawei Feng}
\ead{davyfeng.c@qq.com}

\author[label3]{Xing Zhou}
\ead{zhouxing@nudt.edu.cn}

\author[label4]{Lianghao Li}
\ead{lianghao93@nudt.edu.cn}

\author[label2]{Sheng Qi}
\ead{qisheng@nudt.edu.cn}

\author[label1]{Bo Ding}
\ead{dingbo@nudt.edu.cn}

\author[label1]{Yijie Wang \corref{cor1}}
\ead{wangyijie@nudt.edu.cn}

\author[label2]{Rui Wang}
\ead{ruiwangnudt@nudt.edu.cn}

\author[label1]{Huaimin Wang}
\ead{hmwang@nudt.edu.cn}

\cortext[cor1]{Corresponding author.}

\address[label1]{National Key Laboratory of Parallel and Distributed Computing, College of Computer Science and Technology, National University of Defense Technology, Changsha 410000, P.R. China}
\address[label2]{College of System Engineering, National University of Defense Technology, Changsha 410000, P.R. China}
\address[label3]{College of Intelligence Science and Technology, National University of Defense Technology, Changsha 410073, P.R. China}
\address[label4]{State Key Laboratory of Complex \& Critical Software Environment, College of Information and Communication, National University of Defense Technology, Wuhan 430019, P.R. China}
\address[label5]{National University of Defense Technology, Changsha 410000, P.R. China}

\begin{abstract}
    Existing evolutionary algorithms for Constrained Multi-objective Optimization Problems (CMOPs) typically treat all constraints uniformly, overlooking their distinct geometric relationships with the true Constrained Pareto Front (CPF). In reality, constraints play different roles: some directly shape the final CPF, some create infeasible obstacles, while others are irrelevant. To exploit this insight, we propose a novel algorithm named RCCMO, which sequentially performs unconstrained exploration, single-constraint exploitation, and full-constraint refinement. The core innovation of RCCMO lies in a constraint prioritization method derived from these geometric insights, seamlessly coupled with a unique dual-directional search mechanism. Specifically, RCCMO first prioritizes constraints that constitute the final CPF, approaching them from the evolutionary direction (optimizing objectives) to locate the CPF directly shaped by single-constraint boundaries. Subsequently, for constraints that merely hinder the population's progress, RCCMO searches from the anti-evolutionary direction (targeting the infeasible boundaries where hindering constraints intersect with the CPF) to effectively discover how these constraints obstruct and form the final CPF. Meanwhile, irrelevant constraints are intentionally bypassed. Furthermore, a series of specialized mechanisms are proposed to accelerate the algorithm's execution, reduce heuristic misjudgments, and dynamically adjust search directions in real time. Extensive experiments on 5 benchmark test suites and 29 real-world CMOPs demonstrate that RCCMO significantly outperforms seven state-of-the-art algorithms.
\end{abstract}


\end{frontmatter}

\section{Introduction}
\label{introduction}

Many real-world optimization problems often contain multiple conflicting optimization objectives and must simultaneously satisfy multiple constraints, such as resource scheduling problems\cite{schd}, molecular generation problems\cite{mg}, and Path planning problems\cite{zx1}\cite{zx2}. This type of problem is commonly called Constrained Multi-Objective Optimization Problems (CMOPs). Without loss of generality, a CMOP can usually be defined as follows\cite{CCMO}:

	\begin{flalign}
		\left\{
		\begin{array}{l}
			min\;F(X)=(f_1(X),f_2(X),\ldots,f_m(X)), \\
			subject\;to\;X\; \in \;\Omega ,\\
			\ \ \ \ \ \ \ \ \ \ \ \ \ \ g_i(X)\leq0,i=1,\dots,p\\
			\ \ \ \ \ \ \ \ \ \ \ \ \ \ h_j(X) = 0,j=1,\dots,q
			\label{CMOPs}
		\end{array}
		\right.
	\end{flalign}
    
        where $X=(x_1,x_2,\ldots,x_d)$ represents a $d$-dimensional decision variable vector from the decision space $\Omega$, $F(X)$ denotes an objective function vector consisting of $m$ conflicting objective functions, while $g_i(X)$ refers to $q$ inequality constraints and $h_i(X)$ signifies $p$ equality constraints.
	
	The ultimate goal of solving a CMOP is to obtain a set of feasible solutions, which is the constraint Pareto Front (CPF), that are such that no feasible solution exists which is simultaneously superior to or equal to (with at least one being superior) all solutions in this set across every optimization objective.
    In CMOPs, there are not only feasible solutions but also infeasible solutions in which the satisfaction of each constraint can be measured by $c(X)$ \cite{top}:
	\begin{equation}
		\label{cjx}
		c_j(X)=
		\left\{
		\begin{array}{l}
			max(0,g_i(X)), \ \ \ \ \ \ \ \ \ \ i=1,\ldots,p \\
			max(0,|h_j(X)|, \ \ \ j=p+1,\ldots,p+q 
		\end{array}
		\right.
	\end{equation}
	 By aggregating the degrees of violation for each constraint, represented as $CV(X)$ in equation \ref{CVX}, we ascertain the overall constraint violation of a solution. A solution is deemed feasible if $CV(X)$ equals zero; otherwise, it is regarded as infeasible.
	
	\begin{eqnarray}\label{equation1}
		\label{CVX}
		CV(X) =  \sum_{j=1}^{p+q} c_j(X),            
	\end{eqnarray}

Multi-Objective Evolutionary Algorithms (MOEAs) have proven highly effective in solving standard multi-objective optimization problems. When practical constraints are introduced, Constraint Handling Techniques (CHTs) are typically integrated into MOEAs to form Constrained Multi-Objective Evolutionary Algorithms (CMOEAs). The primary goal of a CMOEA is to simultaneously balance convergence, diversity, and feasibility within a given computational budget. However, the majority of existing CMOEAs directly apply CHTs to the aggregated overall constraint violation, denoted as $CV(X)$. This aggregate approach inherently treats all constraints uniformly, ignoring the topological differences and potential priority relationships among individual constraints. 

Beyond simply overlooking these topological differences, monolithic aggregation fundamentally destroys the multidimensional geometric information of the problem, leading to two critical flaws in complex environments. First, \textbf{scale imbalance} severely blinds the search direction. In real-world applications, constraints often possess vastly heterogeneous physical units and numerical magnitudes (e.g., an elasticity modulus limit of $10^6$ versus a deflection limit of $10^{-3}$). Direct aggregation causes constraints with massive numerical magnitudes to completely overshadow subtle yet geometrically critical constraints. Second, monolithic aggregation creates a \textbf{rugged and deceptive CV landscape}. When diverse, non-linear constraint boundaries are forcibly summed, their overlapping effects artificially generate severe local optima, sharp ridges, and deceptive valleys. Driven by this distorted $CV(X)$ landscape, populations often drift aimlessly in deep infeasible space or get trapped in deceptive local minima, completely missing the narrow pathways leading to the true feasible regions.

To circumvent these geometric blind spots, evaluating and handling constraints independently has emerged as a highly promising paradigm. By isolating individual constraint violations $c(X)$, an algorithm can bypass the distorted aggregate landscape and directly perceive the exact geometric boundary of each specific constraint. This structural isolation makes it significantly easier to identify the specific bottleneck hindering the population, thereby guiding the search smoothly toward isolated feasible regions. 

Recognizing the distinct advantages of this independent constraint handling, a few pioneering methods have attempted to prioritize constraints or utilize their interrelationships. However, while theoretically promising, determining the appropriate processing sequence and priority of constraints remains a significant challenge, and existing approaches exhibit notable limitations due to their reliance on static proxies. For instance, C3M \cite{c3m} determines priority based on the quality of the Single Constraint Pareto Front (SCPF) formed by each individual constraint, assigning higher priority to constraints with worse SCPFs. However, the final CPF is not always directly related to the SCPF of a single constraint (as analyzed in detail below), causing this method to misjudge priorities when constraints intersect in complex ways. MSCMO \cite{MSCMO} ranks constraints based on their infeasibility rates on the Unconstrained Pareto Front (UPF). While intuitive, this heuristic fails to discern priorities when all constraints are completely infeasible on the UPF, rendering them indistinguishable. DPCPRA \cite{DPCPRA} extends the UPF-based ranking of MSCMO by incorporating dynamic resource allocation across an additional population, but it inherently suffers from the same indistinguishability issue under severe UPF infeasibility.

Other approaches attempt to leverage constraint relationships without explicit prioritization, which often leads to severe computational inefficiency or misdirection. MCCMO \cite{mccmo} attempts to gradually merge constraint handling by evaluating the relationships among SCPFs. MTOTC \cite{mtotc} hypothesizes that the final CPF can be approached by relaxing one constraint at a time, maintaining $p+q+1$ distinct populations (where $p+q$ represents the total number of constraints) to relax each constraint simultaneously. Fundamentally, to analyze and utilize the evolving interrelationships among constraints, these cooperative or multi-population methods must continuously update all sub-populations in every single generation to preserve the constraint information. This simultaneous, real-time update requirement constitutes a major structural flaw. It incurs massive non-evaluative computational overhead, such as redundant non-dominated sorting and environmental selections, rendering these algorithms inherently slow and poorly scalable as the number of constraints increases. 

To explicitly illustrate why the final CPF is not always directly related to an individual SCPF, we categorize the geometric relationships between constraints and the final CPF into three distinct topological scenarios, as depicted in Fig. \ref{type}:

\begin{figure*}[htbp]
    \centering
		\subfigure[]{
			\includegraphics[width=5.3cm]{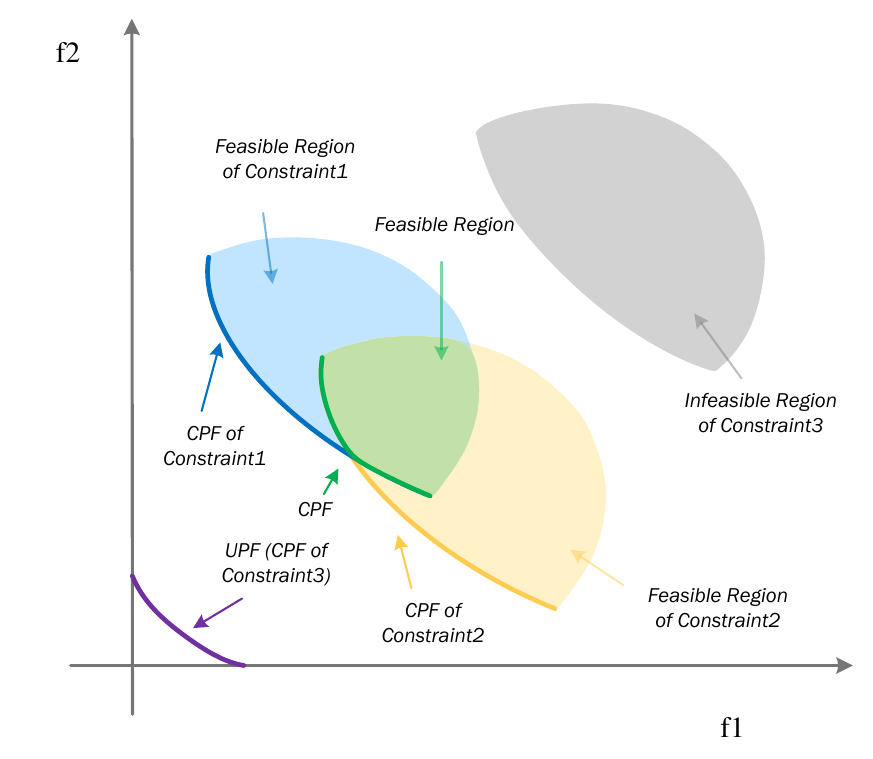}
			\label{type1}
		}
		\subfigure[]{
			\includegraphics[width=5.3cm]{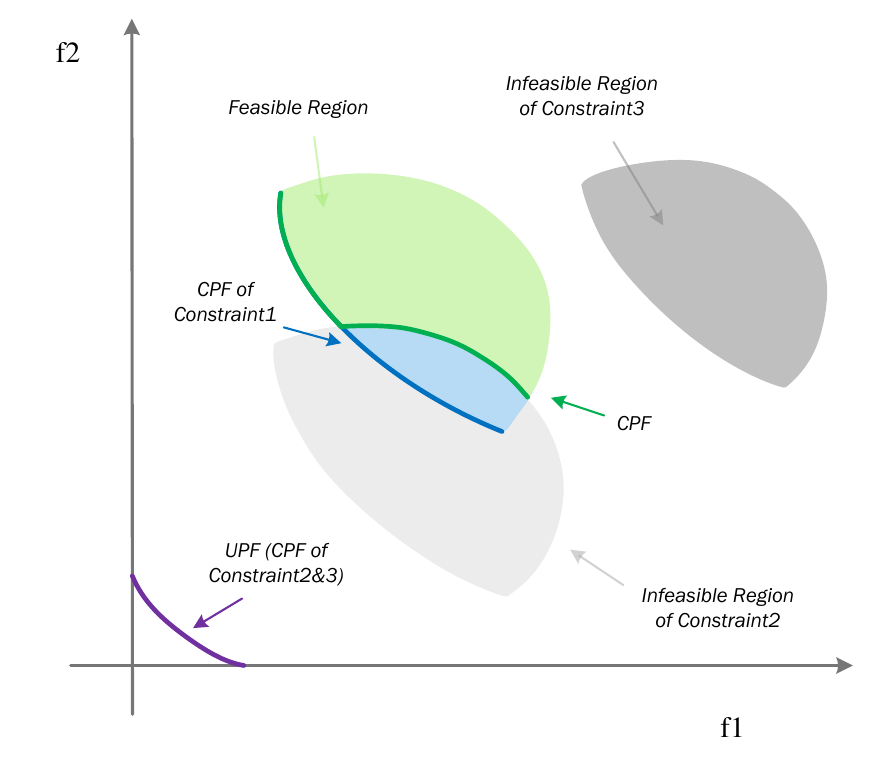}
			\label{type2}
		}
        \subfigure[]{
			\includegraphics[width=5.3cm]{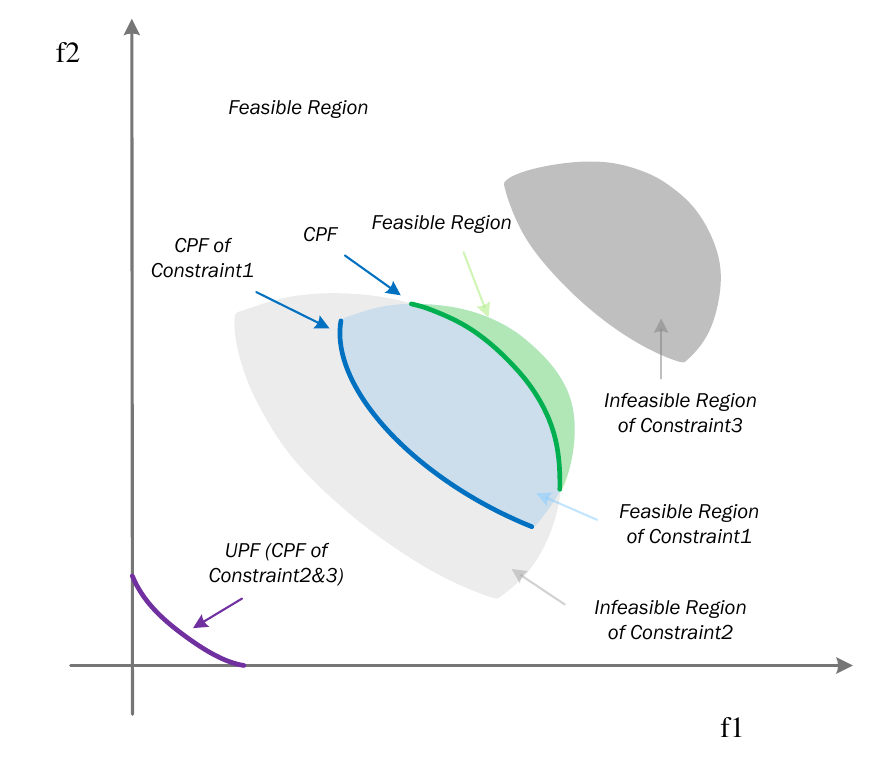}
			\label{type3}
		}
		\caption{Three types of constraint priority situations based on their geometric relationships with the final CPF.}
		\label{type}
\end{figure*}

\begin{enumerate}
    \item The final CPF is exclusively composed of segments from certain SCPFs. In this scenario, constraints whose SCPFs directly constitute the final CPF hold the highest priority, as solving them naturally leads to the final optimum. Constraints that do not intersect or shape the final CPF are deemed irrelevant and assigned the lowest priority. As shown in Fig. \ref{type1}, constraints 1 and 2 have the highest priority to locate their respective SCPFs, whereas constraint 3 is irrelevant.
    
    \item The final CPF is shaped by a combination of certain SCPFs and the intersecting infeasible boundaries of other constraints. Specifically, the infeasible regions of some constraints overlap with the underlying SCPFs, meaning the final feasible region is strictly restricted by these overlapping infeasible borders. Consequently, constraints whose SCPFs form the base of the final CPF receive the highest priority. Constraints whose infeasible boundaries block the search path are assigned a medium priority, as their boundaries must be explicitly located to define the truncated CPF. Irrelevant constraints remain at the lowest priority. As illustrated in Fig. \ref{type2}, constraint 1 holds the highest priority, constraint 2 holds a medium priority, and constraint 3 has the lowest priority.
    
    \item The final CPF is entirely defined by the overlapping infeasible boundaries of various constraints, with no direct contribution from any individual SCPF. In this scenario, every constraint's SCPF coincidentally falls completely within the infeasible regions of other constraints. Because it is impossible to solve the CMOP by merely locating individual SCPFs, constraints whose infeasible boundaries block the search to directly form the final CPF are elevated to the highest priority, while all other constraints are assigned the lowest priority. As shown in Fig. \ref{type3}, constraint 2 has the highest priority to map its infeasible boundary, while constraints 1 and 3 are assigned the lowest priority.
\end{enumerate}

Motivated by these geometric distinctions, we propose a novel algorithm, RCCMO, which is fundamentally a dynamic constraint prioritization method. It implements a structured progression from unconstrained exploration to targeted single-constraint exploitation, and finally to global feasible refinement. To accurately assess the geometric influence of individual constraints, RCCMO assigns a dual-population architecture to each constraint: one population dedicated strictly to the evolutionary search (locating the constraint-specific SCPF), and another to the anti-evolutionary search (approaching the CPF from the outside to map obstructive infeasible boundaries). Simultaneously, a specialized probe population is maintained to continuously detect constraints that actively hinder the evolutionary progress. By dynamically gathering pure feasibility rates from the evolutionary populations and blockage rates from the probe population, RCCMO categorizes and ranks all constraints. The computational effort is then strategically directed to the highest-priority constraint from the appropriate direction, while irrelevant constraints are intentionally bypassed.

However, inferring the exact topological roles of constraints from population statistics is inherently heuristic. To mitigate the risk of premature misclassification, RCCMO incorporates a dynamic priority re-evaluation and a real-time directional correction mechanism. Rather than relying on a static, one-time decision, the algorithm cyclically re-assesses constraint priorities using progressively converging populations. Furthermore, since the initial judgment of the search direction (evolutionary or anti-evolutionary) may occasionally be inaccurate, the real-time correction mechanism is designed to dynamically rectify such misjudgments and ensure the correct search trajectory. Meanwhile, maintaining multiple populations for each constraint inevitably incurs massive non-evaluative computational overhead, which is primarily dominated by redundant environmental selections across generations. To overcome this computational bottleneck, an Asymmetric Update Strategy (AUS) is proposed. By updating populations asymmetrically and periodically based on their active status, AUS successfully eliminates prohibitive computational costs, keeping the algorithm exceptionally fast. Finally, Stage 3 initiates a full-constraint refinement phase, applying a feasibility-priority strategy to converge upon a set of high-quality solutions.

The main contributions of this paper are summarized as follows:

\begin{enumerate}
    \item A novel dynamic constraint prioritization and dual-directional search mechanism is proposed. By independently assessing the geometric role of each constraint, it explicitly categorizes them into three priority levels: CPF-shaping, search-obstructing, and irrelevant constraints. Based on these priorities, constraints are processed from either an evolutionary or anti-evolutionary direction to efficiently outline the exact topology of the constrained Pareto front (CPF).
    
    \item Based on the proposed mechanism, a highly efficient three-stage CMOEA named RCCMO is developed. To ensure topological robustness against heuristic misclassifications, the framework incorporates dynamic priority re-evaluation and real-time directional correction. Furthermore, an Asymmetric Update Strategy (AUS) is introduced to eliminate the massive non-evaluative computational overhead typically associated with multi-population architectures.
    
    \item Extensive experiments on 63 benchmark instances across five diverse test suites and 29 real-world constrained optimization problems demonstrate the superiority of the proposed algorithm. Supported by rigorous statistical analyses, RCCMO achieves state-of-the-art optimization performance and exceptional execution efficiency compared to seven contemporary CMOEAs.
\end{enumerate}

The remainder of this paper is organized as follows. Section II reviews related works. Section III details the proposed RCCMO framework and its core geometric mechanisms. Section IV presents the experimental setup, performance comparisons, and an in-depth mechanism-level diagnosis. Finally, Sections V and VI conclude the paper and discuss future research directions.

\section{Related Works}

Based on the evolutionary phases and the number of concurrent optimization formulations (often referred to as tasks) employed, existing CMOEAs can be broadly classified into four categories:

\subsection{Single-Stage, Single-Task CMOEAs}
The first category comprises single-stage, single-task CMOEAs, which pursue a fixed optimization goal throughout the entire evolutionary process. While early methods in this category often struggled to solve highly complex CMOPs, their core mechanisms remain foundational to modern algorithms. A classic approach is the Constraint Dominance Principle (CDP) introduced with NSGA-II \cite{nsga2}. This framework strictly prioritizes feasibility by favoring solutions with a smaller constraint violation, $CV(X)$, when at least one solution is infeasible. The Self-adaptive Penalty Function (SPF) \cite{spf1} transforms CMOPs into unconstrained MOPs by penalizing objective values with $CV(X)$, though its performance is highly sensitive to the choice of penalty factors. Stochastic Ranking (SR) \cite{sr} balances objectives and constraints by using a probability parameter $P$ to determine whether to compare objective values or $CV(X)$ during individual sorting. The multiobjective-based method (MOB) \cite{mob} treats $CV(X)$ as an additional, independent objective to maintain diverse selection pressure. Recently, this category has seen renewed innovations in specialized search operators. For instance, the PIC algorithm \cite{PIC_2025} introduces a novel population image convolution method to efficiently locate and thoroughly search the feasible region. Similarly, SS-MOEA \cite{SSMOEA_2025} targets large-scale CMOPs by initially focusing on a low-dimensional subspace of high-contribution variables to accelerate early convergence, smoothly expanding to the full decision space without explicit stage divisions.

\subsection{Multi-Stage, Single-Task CMOEAs}
The second category encompasses multi-stage, single-task CMOEAs, which dynamically adjust their optimization strategies across different evolutionary phases. A widely adopted technique is the $\epsilon$-constraint method \cite{ep1}, which relaxes constraint boundaries to treat slightly infeasible solutions as pseudo-feasible at different stages. Building on this, the Push and Pull Search (PPS) \cite{pps} utilizes the $\epsilon$-method to first push the population toward the UPF for broad exploration, and subsequently pull it back to the CPF by tightening $\epsilon$. Similarly, the MOEA/D-DAE \cite{lir} framework employs $\epsilon$-relaxation to help populations escape local optima. Other phase-shifting methods include ToP \cite{top}, which initially transforms the CMOP into a Constrained Single-Objective Optimization Problem (CSOP) to resolve objective conflicts before reverting to multi-objective optimization. CMOEA-MS \cite{cmoeams} shifts its priority from objectives to constraints based on a feasibility ratio threshold $\lambda$. TSTI \cite{tsti} proposes a two-stage, three-indicator framework focusing on diversity in the first stage and rapid CPF convergence in the second. TSCSO \cite{tscso} and CMOES \cite{cmoes} also explicitly separate unconstrained exploration from strict constraint satisfaction across distinct stages. Continuing this trend, recent advancements include CMOEA-TA \cite{CMOEATA_2025}, a two-stage archiving framework that initially encourages exploration through proportion-based constraint relaxation before enforcing strict dominance. Additionally, CP-TSEA \cite{CPTSEA_2025} utilizes an adaptive $\epsilon$-constraint boundary learning mechanism in its competitive first stage to effectively mitigate local optima before transitioning to a strict convergence phase.

\subsection{Single-Stage, Multi-Task CMOEAs}
The third category consists of single-stage, multi-task CMOEAs. These methods concurrently maintain auxiliary optimization formulations (or populations) that exchange information to assist the main constrained optimization process. The seminal CCMO \cite{CCMO} maintains a main population constrained by the original problem and an unconstrained auxiliary population, sharing information via a joint offspring pool. Similar collaborative structures are seen in CTAEA \cite{ctaea}, CMOEA-TCP \cite{cmoeatcp}, and EMCMO \cite{emcmo}, which leverage auxiliary populations to explore unconstrained regions or preserve diversity. DBC-CMOEA \cite{dbccmoea} and CMOSMA \cite{cmosma} facilitate complementarity through archive management and self-organizing maps, respectively. BiCo \cite{bico} and cDPEA \cite{cdpea} approximate the CPF by bridging the gap between the UPF and CPF from opposite evolutionary directions. In CMOEAPP \cite{cmoeapp} and CMAOO \cite{cmaoo}, convergence and diversity are balanced by temporarily ignoring constraints or optimizing additional spatial objectives. Recently, CMOEA-CD \cite{liu2025constraint} proposed a constraint-Pareto dominance relationship within a three-task architecture, while COEA-DAS \cite{liu2024coevolutionary} designed specific mechanisms based on UPF-CPF relationships. Furthermore, MTMOEA \cite{MTMOEA_2025} concurrently optimizes the original task alongside two auxiliary tasks (fully unconstrained and partially constrained) to provide continuous supplementary search directions.

\subsection{Multi-Stage, Multi-Task CMOEAs}
The final category, multi-stage multi-task CMOEAs, integrates phase-based strategic shifts with collaborative parallel populations to handle the most formidable CMOPs. DD-CMOEA \cite{ddcmoea} explicitly splits the evolutionary process into an exploration stage and an exploitation stage. TriP \cite{trip} extends the CCMO concept by adding a third task dedicated to a relaxed CMOP. MSCEA \cite{mscea} tackles constraint-centered subproblems using progressively narrowing boundaries, while DBEMTO \cite{dbemto} incorporates adaptive policy selection to balance intra- and inter-population knowledge transfer. MTCMO \cite{mtcmo} utilizes dynamic auxiliary tasks with adjustable boundaries to help populations overcome severe infeasibilities. CMOEMT \cite{cmoemt}, CMOQLMT \cite{cmoqlmt}, and MOEA-CMT \cite{chu2024competitive} employ sophisticated transitional stages, reinforcement learning, and subtask competition to dynamically allocate resources. Representing the latest developments in this category, CIDEMT \cite{CIDEMT_2026} introduces a two-stage, tri-task framework that adapts knowledge transfer strategies based on a novel constraint-intensity classification, explicitly measuring the overlap between the UPF and CPF. Finally, CCMT \cite{CCMT_2025} models the CMOP using three distinct formulations—constraint-first, constraint-ignored, and constraint-relaxed—that seamlessly collaborate and compete within a two-stage evolutionary process to achieve highly robust optimization.

	\section{PROPOSED ALGORITHM}
    
\subsection{Framework of RCCMO}

\begin{algorithm}[!htbp]
    \scriptsize
    \caption{Framework of the proposed RCCMO}
    \label{A1}
    \KwIn{$N$ (Population size), $Nc$ (Number of constraints), $V$ (Interval), $\beta$ (Budget threshold)}
    \KwOut{$P_0$ (Final population)} 
    
    $P_0 \gets RandomInitialization(N)$ \tcp*{Initialize main population}
    \For{$i = 1 : Nc$}{
        $P_i^{pos} \gets P_0$, $P_i^{neg} \gets P_0$ \tcp*{Initialize dual populations}
        $Dir_i \gets \text{Positive}$ \tcp*{Initialize search directions}
    }
    $P_{Nc+1} \gets P_0$ \tcp*{Initialize UPF population}
        
    $Pc \gets Nc+1$, $R \gets \emptyset$, $U \gets \emptyset$ \tcp*{Initialize states}
    
    \BlankLine
    \While{termination criterion not fulfilled}{
        $O \gets \text{Generate } N \text{ solutions based on } P_{act} \text{ by DE}$\;
        $Pb \gets UpdateProbe(Pb, O)$ \tcp*{Update probe via Alg. \ref{A2}}
        $P_0 \gets EnvSelection(P_0 \cup O, CV)$ \tcp*{Always update main pop}
            
        \BlankLine
        \uIf{$Pc == Nc+1$}{
            \tcp{Stage 1: UPF Exploration}
            $P_{Nc+1} \gets EnvSelection(P_{Nc+1} \cup O, ObjValue)$\;
            \If{generation $\pmod V == 0$}{
                Update all $P_{1:Nc}^{pos}$ and $P_{1:Nc}^{neg}$ via Alg. \ref{A3}\;
            }
            \If{Satisfy the stage transition condition}{ 
                $[Pc, Dir_{Pc}, U, R] \gets DetermineTarget(P_{1:Nc}^{pos}, Pb, U, Nc)$ via Alg. \ref{A4}\;
            }
        }
            
        \BlankLine
        \uElseIf{$0 < Pc \leq Nc$}{
            \tcp{Stage 2: Single-constraint Exploitation}
            \eIf{generation $\pmod V == 0$}{
                Update all $P_{1:Nc}^{pos}$ and $P_{1:Nc}^{neg}$ via Alg. \ref{A3} \tcp*{AUS: Interval update}
            }{
                Update $P_{Pc}^{pos}$ via Alg. \ref{A3} (Positive) \tcp*{AUS: Update active pos}
                \If{$Dir_{Pc} == \text{Negative}$}{
                    Update $P_{Pc}^{neg}$ via Alg. \ref{A3} (Negative) \tcp*{AUS: Update active neg}
                }
            }    

            \uIf{$Dir_{Pc} == \text{Positive} \land \min(CV(P_{Pc}^{pos})) > 0$}{
                $Dir_{Pc} \gets \text{Negative}$ \tcp*{Instant Flip: Map boundary}
            }
            \ElseIf{$Dir_{Pc} == \text{Negative} \land \min(CV(P_{Pc}^{pos})) == 0$}{
                $Dir_{Pc} \gets \text{Positive}$ \tcp*{Instant Flip: Exploit SCPF}
            }

            \If{Satisfy the stage transition condition}{ 
                $[Pc, Dir_{Pc}, U, R] \gets DetermineTarget(P_{1:Nc}^{pos}, Pb, U, Nc)$ via Alg. \ref{A4}\;
            }
            
            \If{FE $> maxFE \times \beta$}{
                $Pc \gets 0$ \tcp*{Trigger Stage 3 based on budget}
            }
        }
        
        \BlankLine
        \ElseIf{$Pc == 0$}{
            \text{Continue evolution exclusively on } $P_0$ \tcp*{Stage 3: CPF Refinement}
        }
    }
          
    \Return $P_0$
\end{algorithm}

\begin{figure*}[!htbp]
	\setlength{\abovecaptionskip}{0pt}
	\setlength{\belowcaptionskip}{0pt}
	\centering
	\includegraphics[width=\textwidth]{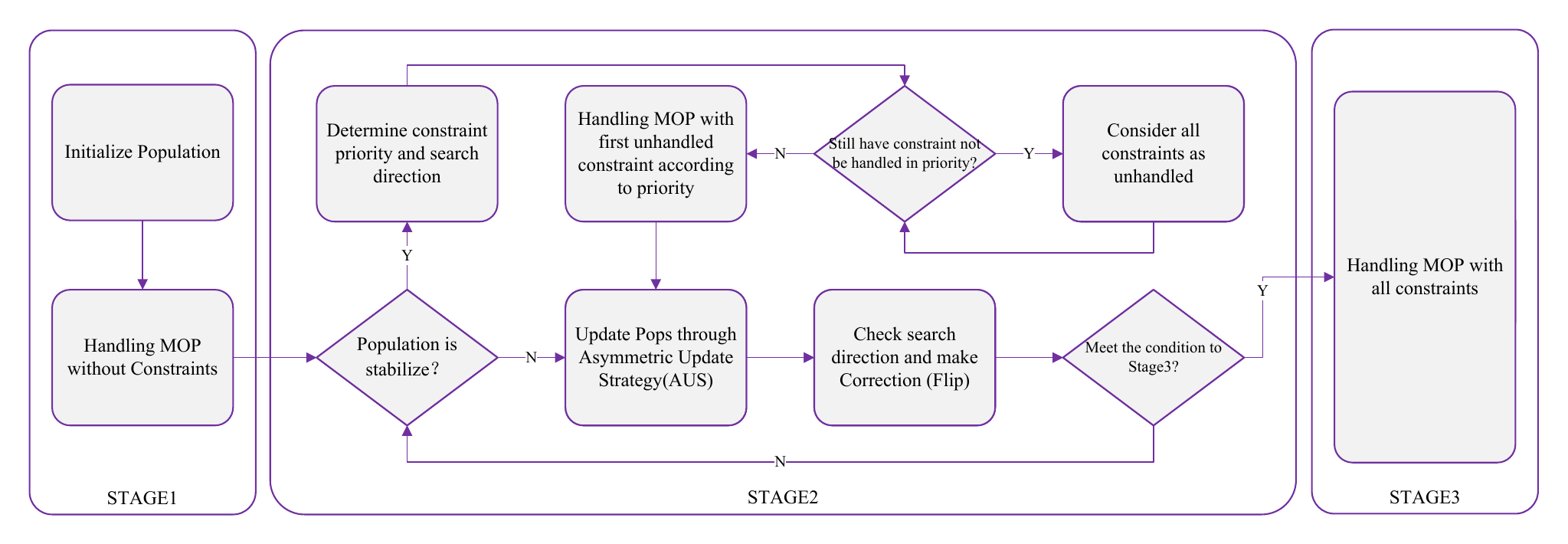}
	\caption{Flowchart of the proposed RCCMO}
	\label{flowchat}
\end{figure*}

The overall procedure of RCCMO is summarized in Algorithm \ref{A1} and illustrated in Fig. \ref{flowchat}. Structurally, the algorithm progresses through three distinct phases: unconstrained exploration (Stage 1), targeted single-constraint exploitation (Stage 2), and global feasible refinement (Stage 3). To accurately assess the distinct geometric roles of individual constraints, RCCMO assigns a dual-population architecture to each of the $Nc$ constraints. Specifically, for the $i$-th constraint, a positive population ($P_i^{pos}$) is strictly dedicated to the evolutionary search (locating the constraint-specific SCPF), while a negative population ($P_i^{neg}$) is dedicated to the anti-evolutionary search (mapping the obstructive infeasible boundary). An additional population, $P_{Nc+1}$, is reserved to explore the Unconstrained Pareto Front (UPF). 

The operational phase and the current target of the algorithm are strictly controlled by the state variable $Pc$. $Pc$ acts as a dynamic indicator: when $Pc = Nc+1$, the algorithm operates in Stage 1 targeting the UPF; when $0 < Pc \leq Nc$, it executes Stage 2, heavily focusing on the $Pc$-th constraint; and when $Pc = 0$, it enters Stage 3 to refine the final Constrained Pareto Front (CPF). Two sets, $R$ and $U$, are maintained to track the dynamic priority ranking and the constraints that have already been processed, respectively.

In each generation, the active population ($P_{act}$) is dynamically selected based on the current stage $Pc$ and its assigned search direction $Dir_{Pc}$. The reproduction operation generates $N$ offspring solutions ($O$) from $P_{act}$ using Differential Evolution (DE) operators. Following reproduction, a specialized probe population, $Pb$, is updated to continuously detect constraints that actively hinder the search progress. Notably, regardless of the current stage, the main population $P_0$ unconditionally executes environmental selection using the newly generated offspring. This ensures that $P_0$ acts as a global archive, immediately absorbing and preserving any high-quality feasible solutions discovered during the exploration of various boundaries.

When operating in Stage 1 ($Pc = Nc+1$), RCCMO temporarily ignores constraint satisfaction to broadly explore the UPF via $P_{Nc+1}$. The UPF serves as a critical baseline for assessing constraint topologies. Stage 1 concludes when the inter-generational objective variations of the population stabilize, employing the same quantitative stability criterion proposed in \cite{c3m}. Once stability is achieved, Algorithm \ref{A4} (`DetermineTarget`) evaluates the initial priorities, outputs the highest-priority constraint to $Pc$ along with its optimal search direction $Dir_{Pc}$, and transitions the algorithm into Stage 2.

Stage 2 ($0 < Pc \leq Nc$) marks the core single-constraint exploitation phase. A fundamental challenge in this stage is maintaining accurate topological information across the evolving search space. This continuous maintenance serves two critical purposes. First, because constraint priorities dynamically shift during exploration, the algorithm must continuously gather fresh statistical data—specifically, the pure feasibility rates stored in the positive populations ($P^{pos}$) and the blockage rates captured by the probe population ($Pb$)—for subsequent priority re-evaluations. Second, as detailed in the subsequent paragraph, this up-to-date topological information is strictly required to enable the real-time directional flipping mechanism. Ideally, to preserve absolute geometric sensitivity, all populations should be updated at every generation. However, updating all $2Nc+1$ populations in every single generation incurs massive non-evaluative computational overhead (e.g., redundant non-dominated sorting and crowding distance calculations), creating a severe execution bottleneck.

We recognized that the macroscopic topological relationships of inactive constraints do not fluctuate drastically across adjacent generations. Therefore, RCCMO employs an Asymmetric Update Strategy (AUS), where all inactive populations are updated only periodically every $V$ generations ($V=30$ in this paper). This preserves the essential topological information with minimal computational cost.

Crucially, for the active constraint $Pc$, AUS is tightly coupled with the Instant Bi-directional Flipping mechanism. Evaluations are allocated highly asymmetrically: $P_{Pc}^{pos}$ is updated every single generation regardless of the current search direction, whereas $P_{Pc}^{neg}$ is updated every generation exclusively when the search direction is negative (otherwise, it strictly follows the $V$-generation interval). The deep rationale behind this design is that $P_{Pc}^{pos}$ serves as the sole, real-time geometric trigger for directional correction. 
Specifically, the flipping conditions strictly depend on the feasibility status of $P_{Pc}^{pos}$. If the current direction is negative (mapping an obstacle), but the continuously updated $P_{Pc}^{pos}$ suddenly discovers a global feasible solution, it geometrically proves that the constraint's SCPF actually constitutes a segment of the final CPF. The algorithm instantly flips to the positive direction for direct exploitation. Conversely, if the direction is positive but $P_{Pc}^{pos}$ fails to retain any global feasible solutions, it indicates that the SCPF is global infeasible rather than a part of the CPF, instantly triggering a flip to the negative direction to map the blockage. Because both flipping conditions rely entirely on monitoring the feasibility of $P_{Pc}^{pos}$, it demands continuous, generation-by-generation updates. $P_{Pc}^{neg}$, however, is only functionally useful during anti-evolutionary boundary mapping, safely allowing its non-evaluative overhead to be minimized during positive searches.

Upon reaching the stage transition condition, Algorithm \ref{A4} seamlessly recalculates the priorities and outputs the next target constraint $Pc$. Stage 2 concludes and shifts to Stage 3 once the consumed evaluations exceed a predefined budget threshold ($\beta \times maxFE$, where $\beta = 0.7$ is adopted). Upon entering Stage 3 ($Pc = 0$), RCCMO initiates the full-constraint refinement phase. Generating offspring exclusively from $P_0$, the algorithm applies a strict feasibility-priority refinement strategy to deeply exploit the overlapping feasible regions, converging upon the final set of high-quality solutions.

\subsection{Key functions in RCCMO}
	
	\subsubsection{Update Probe}
\begin{algorithm}[!htbp]
    \small
    \caption{Update Probe Population ($UpdateProbe$)}
    \label{A2}
    \KwIn{$Pb$ (Candidate probe solutions), $P_0$ (Main population), $N$ (Maximum size)}
    \KwOut{$Pb$ (Updated probe population)}
    
    $Pb \gets \{x \in Pb \mid (\exists y \in P_0, x \prec y) \land (\not \exists z \in P_0, z \prec x)\}$ \tcp*{Filter solutions}
    \If{$|Pb| > N$}{
        \ForEach{$x \in Pb$}{
            \eIf{$x$ is feasible}{
                $CV'(x) \gets \infty$ \tcp*{Penalize feasible solutions}
            }{
                $CV'(x) \gets 0$ \tcp*{Prioritize infeasible solutions}
            }
        }
        $FN \gets \text{NDSort}(-Obj_{Pb}, CV'_{Pb})$ \tcp*{Sort by negative objectives and modified CV}
        $D \gets \text{SPEA2Density}(Obj_{Pb})$ \tcp*{Calculate crowding distance via SPEA2}
        $Fitness \gets FN + D$\;
        $Pb \gets \text{Select top } N \text{ solutions with minimum } Fitness$\;
    }
    \Return $Pb$
\end{algorithm}

\begin{figure}[!htbp]
	\setlength{\abovecaptionskip}{0pt}
	\setlength{\belowcaptionskip}{0pt}
	\centering
	\includegraphics[width=8cm]{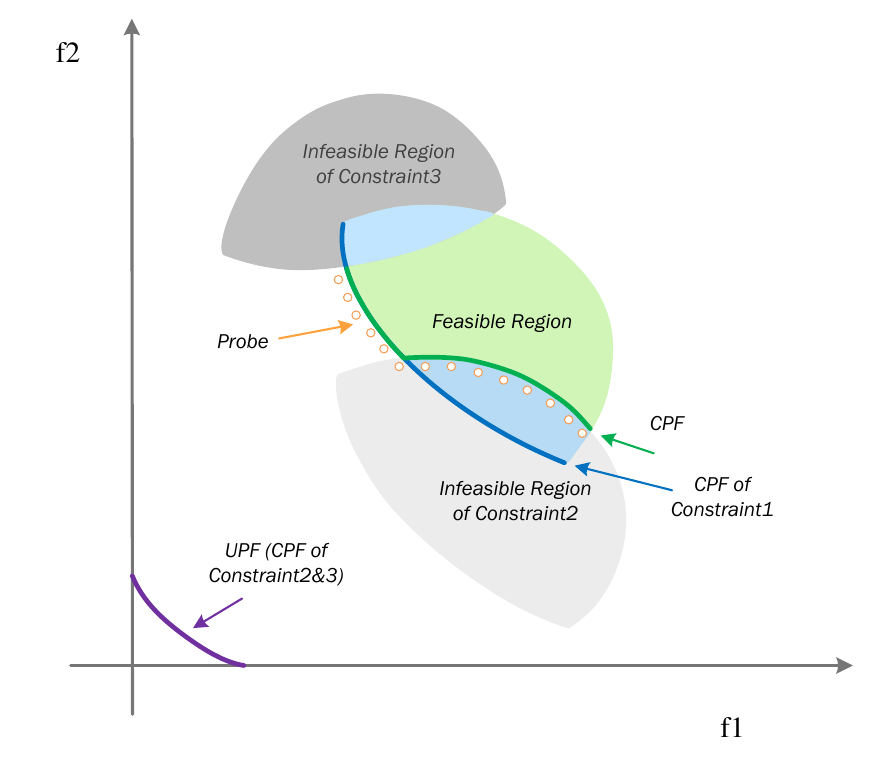}
	\caption{Working principle of Probe population: used to detect constraints that hinder the main population from further evolving towards the Pareto optimal direction.}
	\label{probe}
\end{figure}

The probe population, $Pb$, acts as an exploratory scout to detect the specific constraints that hinder the main population $P_0$ from advancing toward the true Pareto optimal direction. Algorithm \ref{A2} details the update mechanism for this probe population. Initially, the algorithm filters the candidate solutions in $Pb$ by strictly evaluating their dominance relationships with the main population $P_0$. Specifically, it retains only those solutions in $Pb$ that dominate at least one solution in $P_0$ while not being dominated by any solution in $P_0$. This rigorous filtering serves a dual purpose: first, eliminating solutions dominated by $P_0$ prevents the probe from regressing into regions already surpassed by the main search; second, requiring the probe solutions to actively dominate members of $P_0$ ensures they represent the immediate frontier blocking the main population's progress.

If the number of retained solutions exceeds the predefined capacity $N$, a specialized truncation mechanism is activated to isolate the most informative boundary solutions. Because the primary objective of the probe population is to map the obstructive infeasible boundaries rather than feasible regions, feasible solutions in $Pb$ are deliberately penalized by assigning them an infinite constraint violation value, whereas all infeasible solutions are assigned a uniform violation value of zero. A fast non-dominated sorting procedure (NDSort) is then applied using the negative objective values alongside these modified constraint violations. By sorting based on negative objectives, the algorithm actively pushes the selection away from the unconstrained front. Combined with the severe penalty on feasible solutions, this mechanism effectively forces the probe population to accumulate along the extreme edges of the infeasible regions adjacent to $P_0$. 

To maintain a well-distributed representation of these boundaries, the algorithm subsequently calculates the density of each solution using the $k$-th nearest neighbor method derived from SPEA2. The overall fitness of each solution is defined as the sum of its non-dominated front number and density, and the top $N$ solutions with the minimum fitness values are preserved for the next generation. 

To provide a more intuitive geometric understanding, Fig. \ref{probe} explicitly illustrates the dynamic behavior of the probe population. Because the probe solutions (represented by orange circles) are strictly required to dominate the main population $P_0$ (which approximates the current CPF), they are inherently cast into the infeasible space, advancing toward the unconstrained lower-left area (UPF). However, driven by the fast non-dominated sorting based on negative objective values ($-Obj$), these solutions are not allowed to blindly escape deep into the UPF. Instead, minimizing $-Obj$ effectively maximizes the original objectives, acting as a backward repulsive force that continuously pulls the probe solutions back toward $P_0$. 

Sandwiched between the forward dominance requirement and the backward $-Obj$ sorting, and further combined with the mechanism that deliberately penalizes feasible solutions, the probe population is ultimately forced to compactly adhere to the exact intersections where active infeasible boundaries (Constraint 1 and Constraint 2) physically truncate the CPF. In contrast, the infeasible region of Constraint 3 is located away from this immediate convergence trajectory. As visually evident, all probe solutions naturally fall outside the gray infeasible region of Constraint 3, completely satisfying it. Consequently, the infeasibility rate of Constraint 3 within $Pb$ evaluates to strictly zero. Through this geometrically constrained scouting, the probe mechanism successfully isolates the active hindering constraints (Constraints 1 and 2) while implicitly filtering out non-obstructive ones (Constraint 3), thereby establishing a precise, topology-driven foundation for the subsequent priority determination.

\subsubsection{Update Constraint-Specific Populations}

\begin{algorithm}[!htbp]
    \small
    \caption{Interval Update for Dual Auxiliary Populations ($UpdatePopulation$)}
    \label{A3}
    \KwIn{$P_0$ (Main pop), $P_i^{pos}$, $P_i^{neg}$ (Dual populations for constraint $i$), $O$ (Offspring), $N$ (Size)}
    \KwOut{Updated $P_i^{pos}$ and $P_i^{neg}$}
    
    \tcp{1. Update the Positive Population (Evolutionary Search)}
    $C^{pos} \gets P_i^{pos} \cup O$\;
    $P_i^{pos} \gets EnvSelection(C^{pos}, N)$ based solely on constraint $i$'s violation $c_i$\;
    
    \BlankLine
    \tcp{2. Update the Negative Population (Anti-evolutionary Search)}
    $C^{neg} \gets P_i^{neg} \cup O$\;
    $P_i^{vio} \gets \{x \in C^{neg} \mid c_i(x) > 0\}$ \tcp*{Violators of constraint $i$}
    $P_i^{sat} \gets C^{neg} \setminus P_i^{vio}$ \tcp*{Satisfiers of constraint $i$}
    
    \uIf{$|P_i^{vio}| \leq N$}{
        $Fit_{vio} \gets \text{CalculateFitness}(P_i^{vio})$ with overall CV as an extra objective\;
        $Fit_{sat} \gets \text{CalculateFitness}(P_i^{sat}) + \max(Fit_{vio})$ \tcp*{Penalize satisfiers}
        $P_i^{neg} \gets P_i^{vio} \cup \text{Select } (N - |P_i^{vio}|) \text{ solutions from } P_i^{sat} \text{ based on } Fit_{sat}$\;
    }
    \Else{
        $D_{dom} \gets \{x \in P_i^{vio} \mid \exists y \in P_0, x \prec y\}$ \tcp*{Solutions dominated by $P_0$}
        $Fit_{vio} \gets \text{CalculateFitness}(P_i^{vio})$ with overall CV as an extra objective\;
        $\forall x \in D_{dom}, Fit_{vio}(x) \gets Fit_{vio}(x) + \max(Fit_{vio})$ \tcp*{Penalize regression}
        
        $P_{top}^{vio} \gets \{x \in P_i^{vio} \mid Fit_{vio}(x) < 1\}$ \tcp*{Top-tier candidates}
        \eIf{$|P_{top}^{vio}| > N$}{
            $Fit'_{top} \gets \text{CalculateFitness}(P_{top}^{vio})$ using $-Obj$ \tcp*{Anti-evolutionary push}
            $P_i^{neg} \gets \text{Select } N \text{ solutions from } P_{top}^{vio} \text{ based on } Fit'_{top}$\;
        }{
            $P_i^{neg} \gets \text{Select } N \text{ solutions from } P_i^{vio} \text{ based on } Fit_{vio}$\;
        }
    }
    \Return $P_i^{pos}, P_i^{neg}$
\end{algorithm}

Algorithm \ref{A3} details the environmental selection mechanism for updating the dual constraint-specific auxiliary populations. The operations are highly customized to fulfill the distinct geometric missions of $P_i^{pos}$ and $P_i^{neg}$.

The positive population ($P_i^{pos}$) is strictly tasked with finding the specific Single Constraint Pareto Front (SCPF) of the $i$-th constraint. To prevent its trajectory from being distorted by the complex multi-constraint landscape, the environmental selection for $P_i^{pos}$ calculates constraint violations based \textit{solely} on constraint $i$, temporarily ignoring all other constraints. This structural isolation ensures that $P_i^{pos}$ converges purely and accurately toward its target geometric boundary.

Conversely, the negative population ($P_i^{neg}$) executes the anti-evolutionary search to map the obstructive infeasible boundaries. To achieve this, the combined offspring pool is strictly partitioned into a violator set ($P_i^{vio}$, representing solutions that violate constraint $i$) and a satisfier set ($P_i^{sat}$, representing solutions satisfying constraint $i$). Because the explicit goal is to trace the infeasible obstacle, violators are the primary geometric assets, while satisfiers are fundamentally misaligned with this mission.

The subsequent selection logic is driven by the abundance of these violators. If the number of violators is insufficient ($|P_i^{vio}| \leq N$), all violators are unconditionally preserved. To geometrically distinguish them and prevent them from drifting aimlessly into deep infeasible space, their fitness is calculated by treating their \textit{overall} constraint violation as an additional optimization objective. This operation pulls the violators back toward the global feasible region, ensuring they hug the boundary closely. The remaining population slots are filled by satisfiers; however, a severe fitness penalty is applied to these satisfiers to ensure that any newly generated violator in future generations will instantly replace them.

When the number of violators exceeds the population size $N$ ($|P_i^{vio}| > N$), a more rigorous selection is required to determine which infeasible solutions are most helpful for mapping the boundary. This is executed in three concrete steps:
First, the algorithm compares the violators against the main population $P_0$. If a violator is dominated by any solution in $P_0$ (stored in the set $D_{dom}$), it means its objective values are already worse than the currently found feasible (or near-feasible) solutions. Such inferior violators are located far behind the optimal boundary and provide no useful information. Therefore, they are heavily penalized to ensure their elimination.
Second, for the remaining violators, the algorithm calculates their fitness by treating the overall constraint violation as an additional objective. This step is designed to select violators that have good objective values while remaining as close to the feasible region as possible. 
Finally, a critical situation arises if the number of top-tier violators (those in the first non-dominated front, $P_{top}^{vio}$) still exceeds $N$. If the algorithm simply selects them by minimizing their objective values (as standard MOEAs do), all solutions will rapidly crowd into a narrow, overlapping area closest to the unconstrained front. This crowding effect prevents the algorithm from exploring the full length of the constraint boundary. To solve this, RCCMO performs an anti-evolutionary truncation using negative objective values ($-Obj$). Mathematically, minimizing $-Obj$ is equivalent to maximizing the objectives. This counter-intuitive operation deliberately pushes the solutions away from the crowded center, forcing them to spread out along the constraint boundary. By doing so, the algorithm successfully maps the entire boundary instead of just a single point, which is essential for accurately determining how this constraint intersects with the final CPF.

\subsubsection{Determine Target Constraint and Priority}

\begin{algorithm}[!htbp]
    \small
    \caption{Determine Target Constraint ($DetermineTarget$)}
    \label{A4}
    \KwIn{$P_{1:Nc}^{pos}$ (Positive populations), $Pb$ (Probe population), $U$ (Processed set), $Nc$}
    \KwOut{$Pc$ (Target constraint), $Dir_{Pc}$ (Search direction), $U$ (Updated set), $R$ (Priority list)}
    
    \For{$i = 1$ \KwTo $Nc$}{
        \tcp{Gather raw statistical rates}
        $F[i] \gets$ Proportion of solutions in $P_i^{pos}$ where overall $CV(X) == 0$\;
        $IF[i] \gets$ Infeasibility rate of solutions in $Pb$ violating constraint $i$\;
        
        \If{$F[i] > 0$}{
            $IF[i] \gets 0$ \tcp*{Isolate the primary geometric role}
        }
    }
    
    $R_F \gets$ Indices of non-zero elements in $F$, sorted descending\;
    $R_{IF} \gets$ Indices of non-zero elements in $IF$, sorted descending\;
    
    $R \gets R_F \oplus (-R_{IF})$ \tcp*{Irrelevant constraints are strictly excluded}
    
    \BlankLine
    $R' \gets R \setminus U$ \tcp*{Filter out already processed constraints}
    \If{$R' == \emptyset$}{
        $U \gets \emptyset$ \tcp*{Reset the round if all active constraints are processed}
        $R' \gets R$\;
    }
    
    $Pc \gets |R'[1]|$ \tcp*{Extract the absolute index as the target}
    $U \gets U \cup \{Pc\}$ \tcp*{Mark the target as processed}
    $Dir_{Pc} \gets \text{Negative if } R'[1] < 0 \text{ else Positive}$\;
    
    \Return $Pc, Dir_{Pc}, U, R$\;
\end{algorithm}

\begin{figure}[!htbp]
	\setlength{\abovecaptionskip}{0pt}
	\setlength{\belowcaptionskip}{0pt}
	\centering
	\includegraphics[width=\textwidth]{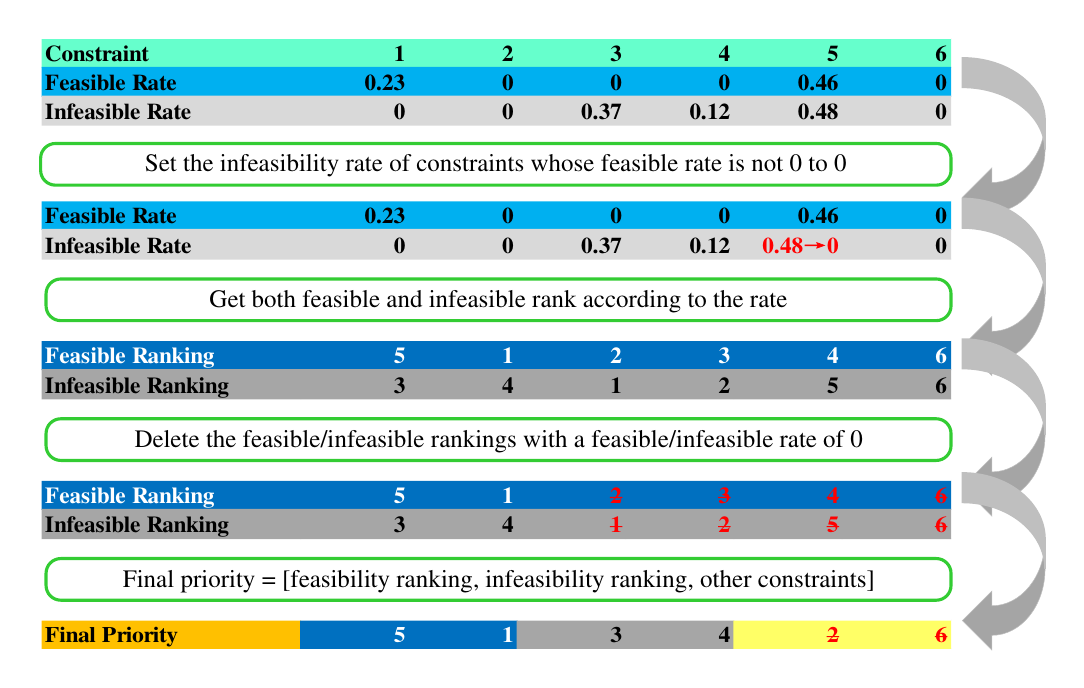}
	\caption{A step-by-step example demonstrating the constraint prioritizing mechanism in RCCMO.}
	\label{priority}
\end{figure}

Algorithm \ref{A4} encapsulates the logical process for determining the comprehensive priority ranking and dynamically outputting the immediate target constraint ($Pc$) along with its optimal search direction ($Dir_{Pc}$). To provide a more intuitive understanding of this sorting mechanism, Fig. \ref{priority} demonstrates a step-by-step numerical example involving six constraints.

The prioritization evaluates two distinct statistical indicators: the global feasibility rate ($F$) extracted strictly from the positive populations ($P^{pos}$), and the probe infeasibility rate ($IF$) captured by the probe population ($Pb$). As shown in the first two rows of Fig. \ref{priority}, Constraint 5 simultaneously exhibits a high feasibility rate (0.46) and a high probe infeasibility rate (0.48). Geometrically, this occurs because a constraint whose boundary directly shapes the final CPF will inevitably also act as a physical obstacle against the unconstrained search path of the probe population. However, within our topological hierarchy, directly contributing to the final CPF is a primary, higher-order property that dictates an evolutionary search direction. To ensure each constraint is uniquely categorized by its most significant geometric role, the algorithm enforces a strict isolation rule. If a constraint has already proven its direct contribution to the CPF ($F[i] > 0$), its secondary obstructive property ($IF[i]$) is intentionally set to 0. As explicitly illustrated in the second step of Fig. \ref{priority}, the original $IF$ value of Constraint 5 (0.48) is cleared to 0 (highlighted in red). This operation ensures that CPF-shaping constraints are ranked exclusively by their feasible contributions, reserving the $IF$ ranking strictly for constraints that merely block the search without providing global feasibility.

Next, the algorithm independently sorts the non-zero elements. Sorting the feasible rates (blue row) yields the highest-priority ranking $R_F = [5, 1]$. Sorting the remaining non-zero infeasible rates (gray row) yields the medium-priority ranking $R_{IF} =[3, 4]$. Constraints 2 and 6, which possess zero rates in both categories, are identified as irrelevant constraints. Because they provide no structural boundaries to the CPF, they are logically treated as already processed and are entirely excluded from the priority list.

The overall active priority list $R$ is constructed by concatenating $R_F$ and $R_{IF}$. To seamlessly communicate the required search direction along with the priority, RCCMO employs a sign-encoding operation. The indices in $R_{IF}$ are multiplied by $-1$ before concatenation. Consequently, the final priority array becomes $R = [5, 1, -3, -4]$. The positive values (5, 1) indicate that these constraints hold the highest priority and must be approached from the positive (evolutionary) direction to exploit their specific CPFs. The negative values (-3, -4) dictate a medium priority, instructing the algorithm to actively map their boundaries from the negative (anti-evolutionary) direction. 

After generating the comprehensive ranking $R$, the algorithm determines the immediate target. It filters out active constraints that have already been processed in the current round (recorded in set $U$). If all active constraints have been exhausted but computational budget remains, $U$ is cleared to initiate a new round, thereby enhancing the algorithm's robustness against early statistical misjudgments. The absolute value of the highest-ranked unprocessed constraint is extracted as the new target $Pc$, and its search direction is assigned based on its sign. These specific parameters are then returned to precisely guide the targeted exploitation in the subsequent generation.

\section{Experimental Studies}
\label{c4}
All experiments in this paper are conducted on the PlatEMO platform \cite{2017PlatEMO}. Unless otherwise specified, the default parameters provided by the platform are utilized.

\subsection{Experimental Settings}
\label{c4.1}
To rigorously evaluate the performance of RCCMO, we selected five challenging benchmark suites featuring multiple constraints—LIRCMOP \cite{lir}, DASCMOP \cite{dascmop}, DOC \cite{top}, SDC \cite{imtcmo}, and ZXH\_CF \cite{zxhcf}—alongside a collection of real-world constrained multi-objective optimization problems (RWMOPs) \cite{rwmop}. Each benchmark suite is specifically designed to assess different algorithmic capabilities under diverse topological difficulties. The LIRCMOP suite features exceptionally large infeasible regions that obstruct the evolutionary path, where the final CPF is often entirely shaped by non-optimal constraint boundaries rather than the unconstrained Pareto front (UPF). The DASCMOP suite introduces specific parameters to explicitly control diversity, feasibility, and convergence difficulties, frequently creating massive feasible islands that are disjoint from the UPF. The DOC suite embeds intricate constraints simultaneously in both the decision and objective spaces, generating a highly oscillatory and deceptive constraint-violation (CV) landscape. The SDC suite emulates real-world complexities by incorporating mixed interactions among high-dimensional constraint functions and coupling scalable distance variables, severely inducing local optima. The ZXH\_CF suite further couples position and distance variables on nonlinear hypersurfaces, presenting formidable topological barriers. Finally, the RWMOPs encompass diverse engineering tasks across mechanical design, chemical processing, and power systems, validating the practical utility of the algorithms under strict multi-physics formulations.

For comparative analysis, seven state-of-the-art CMOEAs were selected based on two primary rationales. First, to demonstrate competitiveness against the latest general advancements in evolutionary computation, we included two recently proposed algorithms: APSEA \cite{apsea}, which features an adaptive population sizing mechanism; and IMTCMO \cite{imtcmo}, which employs angle-based parent selection alongside an $\epsilon$-guided auxiliary population. 

Second, to rigorously validate our specific constraint-handling mechanisms, we selected five highly relevant algorithms designed with similar motivations—namely, leveraging the underlying geometric relationships among multiple constraints through prioritization, decomposition, or structural relaxation. This geometrically motivated group includes MSCMO \cite{MSCMO}, a pioneering method that prioritizes constraints based on unconstrained Pareto front infeasibility rates; C3M \cite{c3m}, a multi-stage baseline that determines priority using dominance relationships among single-constraint CPFs; and MCCMO \cite{mccmo}, a constraint decomposition method that assigns specific sub-populations to each constraint and merges them dynamically. Additionally, we included MTOTC \cite{mtotc}, which explores constraint geometries by maintaining several population replicas to evaluate constraints individually; and FCDS \cite{fcds}, which diverges from traditional aggregated constraint violation ($CV(X)$) methods by explicitly evaluating the number of constraints violated by each solution through a fuzzy constraint dominance strategy.

The parameter configurations for all compared algorithms are strictly set according to their original publications. Across all CMOEAs, the population size $N$ is set to 100, and the maximum number of function evaluations ($FEs$) is fixed at 100,000 for every problem instance. The number of objectives $M$ and decision variables $D$ remain consistent with those defined in their original papers. Comprehensive details regarding the parameter settings can be found in Table ST-I of the supplementary material.

To assess the performance of the algorithms on the benchmark test suites where the true CPF is known, we utilize the Inverted Generational Distance (IGD) \cite{igdm}. The IGD is calculated as shown in Formula \ref{IGD}:
\begin{equation}
    IGD(P,Q) = \frac{\sum_{v\in P}D(v,Q)}{|P|},
    \label{IGD}
\end{equation}
where $P$ is a set of points uniformly distributed along the true CPF, $|P|$ represents the number of points in $P$, $Q$ denotes the optimal solution set obtained by the evaluated algorithm, and $D(v,Q)$ is the minimum Euclidean distance from a point $v$ to the obtained population $Q$. Approximately 10,000 reference points are sampled on the true CPF for the IGD calculation in this study.

For the RWMOPs, where the true CPF is inherently unknown, we employ the Hypervolume (HV) \cite{HV} metric to evaluate performance:
\begin{equation}
    HV = L\left(\bigcup_{i\in Q}[f_1(i),r_1]\times \cdots \times[f_m(i),r_m]\right),
    \label{HV}
\end{equation}
where $L(\cdot)$ denotes the Lebesgue measure, and $R = (r_1, r_2, \dots, r_m)$ is the reference point in the objective space that is strictly dominated by the entire CPF. In this paper, the reference point for calculating HV is uniformly set to $[1, 1, \dots, 1]$ corresponding to the number of objectives $M$. 

To ensure statistical reliability, each algorithm is executed for 30 independent runs on every CMOP instance. The Wilcoxon rank-sum test \cite{llh1} at a 0.05 significance level is utilized to evaluate pairwise statistical differences between RCCMO and the comparative algorithms. Furthermore, to rigorously assess the overall performance rankings across multiple test instances, the Friedman test alongside the Nemenyi post-hoc test with a critical difference (CD) evaluation is conducted.

\subsection{Experimental Results}
\label{c4.2}

\begin{table*}[htbp]
  \centering
  \caption{Statistical Summary of IGD Values ($+$ / $-$ / $\approx$) and Average Ranks across Test Suites.}
  \label{tab:summary_IGD}
  \resizebox{\textwidth}{!}{
  \begin{tabular}{lcccccccc}
    \toprule
    Problem Suite & APSEA & C3M & FCDS & IMTCMO & MCCMO & MSCMO & MTOTC & RCCMO \\
    \midrule
    DASCMOP ($+$/$-$/$\approx$) & 4/5/0 & 1/6/2 & 5/3/1 & 3/3/3 & 0/6/3 & 0/9/0 & 0/7/2 & - \\
    \quad\textit{Avg. Rank} & 5.22 & 5.11 & \textbf{2.67} & 3.67 & 4.89 & 6.22 & 5.11 & 3.11 \\
    \midrule
    DOC ($+$/$-$/$\approx$) & 0/8/1 & 3/6/0 & 0/7/2 & 3/2/4 & 4/4/1 & 0/8/1 & 0/6/3 & - \\
    \quad\textit{Avg. Rank} & 6.22 & 3.67 & 6.22 & 2.67 & \textbf{2.44} & 7.11 & 5.22 & 2.44 \\
    \midrule
    LIRCMOP ($+$/$-$/$\approx$) & 0/14/0 & 0/14/0 & 2/9/3 & 1/11/2 & 1/12/1 & 0/13/1 & 1/9/4 & - \\
    \quad\textit{Avg. Rank} & 7.93 & 5.21 & 3.29 & 4.93 & 3.71 & 5.50 & 3.86 & \textbf{1.57} \\
    \midrule
    SDC ($+$/$-$/$\approx$) & 0/15/0 & 1/14/0 & 2/8/5 & 6/5/4 & 0/15/0 & 0/14/1 & 1/14/0 & - \\
    \quad\textit{Avg. Rank} & 5.27 & 5.33 & 3.93 & 2.33 & 5.80 & 6.13 & 5.33 & \textbf{1.87} \\
    \midrule
    ZXH\_CF ($+$/$-$/$\approx$) & 6/5/5 & 2/14/0 & 6/6/4 & 2/13/1 & 1/15/0 & 0/16/0 & 1/15/0 & - \\
    \quad\textit{Avg. Rank} & 3.69 & 5.56 & 3.44 & 4.06 & 5.00 & 7.38 & 4.69 & \textbf{2.19} \\
    \midrule
    \textbf{Overall} ($+$/$-$/$\approx$) & \textbf{10/47/6} & \textbf{7/54/2} & \textbf{15/33/15} & \textbf{15/34/14} & \textbf{6/52/5} & \textbf{0/60/3} & \textbf{3/51/9} & \textbf{Baseline} \\
    \textbf{Overall Avg. Rank} & \textbf{5.59} & \textbf{5.10} & \textbf{3.81} & \textbf{3.59} & \textbf{4.52} & \textbf{6.46} & \textbf{4.79} & \textbf{2.14} \\
    \bottomrule
    \multicolumn{9}{l}{\textit{Note:} Nemenyi test Critical Difference (CD) = 1.32 at $\alpha=0.05$.}\\
  \end{tabular}
  }
\end{table*}

\subsubsection{Overall Statistical Superiority}
Tables \ref{tab:summary_IGD} summarize the statistical performance and detailed IGD values of RCCMO alongside seven state-of-the-art CMOEAs across 63 benchmark instances. A Friedman non-parametric test across all problem instances yields a p-value strictly less than $10^{-4}$, indicating highly significant performance differences among the algorithms. Consequently, a Nemenyi post-hoc test with a significance level of $\alpha = 0.05$ was conducted, yielding a critical difference (CD) of 1.32. 

RCCMO achieves an exceptional overall average rank of 2.14. Based on the rigorous CD threshold ($2.14 + 1.32 = 3.46$), RCCMO statistically and significantly outperforms all seven compared algorithms, with the closest competitor, IMTCMO, trailing outside the threshold at a rank of 3.59. Notably, RCCMO achieves a commanding win rate, completely dominating traditional prioritization and multi-population methods like MSCMO (60 wins, 0 losses), C3M (54 wins), and MCCMO (52 wins). Furthermore, it maintains a decisive advantage over highly competitive contemporary baselines such as FCDS (33 wins vs. 15 losses) and IMTCMO (34 wins vs. 15 losses).

\subsubsection{Superiority on LIRCMOP and SDC}
The topological mapping capability of RCCMO is most profoundly highlighted on the LIRCMOP and SDC test suites, where it achieves dominant average ranks of 1.57 and 1.87, respectively. The LIRCMOP suite is characterized by exceptionally large infeasible regions that explicitly sever the true CPF, while the SDC suite introduces complex distance variable linkages that induce severe convergence blockages. 

The detailed data reveals exactly why other geometrically motivated CMOEAs failed in these environments. Methods like C3M and MCCMO fundamentally restrict their constraint handling to the positive (evolutionary) search direction. Consequently, they struggle significantly on instances like LIRCMOP1--4, where medium-priority constraints form obstructive infeasible boundaries that must be explicitly mapped from the outside. Furthermore, their initial heuristic inferences are prone to misjudgments, and they critically lack any dynamic error-correction mechanism to recover from them. 

Conversely, RCCMO precisely identifies these massive blocks using its probe population ($Pb$). By isolating the offending constraint into a dedicated negative population ($P^{neg}$) and deliberately worsening objective values, RCCMO strictly traces the exact contour of these rigid boundaries from the anti-evolutionary direction. Paired with the Asymmetric Update Strategy (AUS), RCCMO heavily concentrates its evaluations on this active boundary and employs real-time directional flipping to correct any initial misclassifications. This comprehensively outmaneuvers static, unidirectional methods, yielding IGD improvements often by an order of magnitude.

\subsubsection{Diagnosis on DOC}
On the DOC suite, which embeds intricate constraints directly within the decision space to generate a highly multimodal and deceptive constraint-violation ($CV(X)$) landscape, RCCMO ties for the top position (Average Rank: 2.44). 

Algorithms relying on the unconstrained Pareto front (UPF) for prioritization, such as MSCMO, frequently fail to find any feasible solutions (yielding "NaN" for IGD) on instances like DOC2 and DOC8. This catastrophic failure occurs because the UPF is often located within deep, deceptive infeasible valleys, rendering UPF-based sorting completely inaccurate. Misguided by this inaccurate priority, MSCMO wastes its limited FE budget exploring irrelevant regions and exhaustively depletes its resources without converging to the final CPF. 

In stark contrast, RCCMO successfully navigates these notoriously difficult instances (achieving an IGD of $2.91 \times 10^{-1}$ on DOC2). This breakthrough is attributed to the Instant Bi-directional Flipping mechanism. By continuously maintaining a positive population ($P^{pos}$) in the background, RCCMO instantly detects when a deceptive boundary unexpectedly yields a feasible solution. It immediately halts boundary mapping and flips the direction to positive, allowing the algorithm to seamlessly escape local traps and exploit the newly discovered feasible pockets without squandering its evaluation budget.

\subsubsection{Performance on DASCMOP and ZXH\_CF}
Although RCCMO establishes an absolute statistical superiority overall, a rigorous diagnosis of specific instances highlights localized structural trade-offs within our framework. On the DASCMOP suite, FCDS secures the best average rank (2.67), while RCCMO follows closely (3.11). It is crucial to note that RCCMO does not fail to locate the feasible regions; in fact, empirical observations confirm that RCCMO successfully finds and fully covers the entire global CPF on these instances. The performance gap here strictly pertains to local convergence precision rather than global topological mapping. 

The DASCMOP suite creates massive feasible islands that are geometrically distant and completely disjoint. Because RCCMO fundamentally prioritizes macroscopic structural discovery—dedicating a substantial portion of its budget to learning constraint interrelationships and tracing boundaries from the outside—it occasionally triggers the hard budget threshold ($\beta = 0.7$) before completing microscopic fine-tuning. This macro-level focus leaves a relatively limited evaluation budget for Stage 3 to perform deep local convergence. In such severely disjoint topologies, algorithms utilizing fuzzy constraint dominance (like FCDS) aggressively exploit local pockets to achieve slightly better micro-convergence.

Similarly, on the ZXH\_CF suite, RCCMO ranks first overall (2.19), but APSEA and FCDS occasionally secure localized advantages. This suite embeds complex nonlinear linkages between position and distance variables. Under tight variable linkage, the heuristic priority ranking in RCCMO's early evolutionary stages experiences stochastic noise, causing temporary priority misclassifications. While RCCMO's dynamic re-evaluation mechanism robustly corrects these misclassifications over time, the evaluations expended during this self-correction process delay the final refinement. This slightly compromises its micro-convergence precision on highly linked landscapes. However, as evidenced by the overall statistics, RCCMO's architectural focus on correct global topological mapping yields a far more robust and consistent performance across the broader spectrum of constrained optimization challenges.

\subsubsection{Performance on Real-World Applications (RWMOPs)}
To validate the practical utility of the proposed framework, RCCMO was evaluated on a comprehensive suite of 29 Real-World Multi-Objective Constrained Optimization Problems (RWMOPs). As detailed in the supplementary document, these problems span complex engineering domains, including mechanical design, chemical process synthesis, and power system optimization. Tables \ref{tab:summary_HV}  present the statistical summary and detailed Hypervolume (HV) results. Among the eight evaluated algorithms, RCCMO achieves the best overall average rank of 3.21. According to the Nemenyi test (CD = 1.98), RCCMO establishes a statistically significant advantage over algorithms like MSCMO (5.93) and FCDS (5.32). More importantly, RCCMO maintains a commanding numerical superiority over all competitors, completely dominating methods like C3M and APSEA (both 15 wins vs. 5 losses), while decisively outperforming strong multi-population baselines like MTOTC (13 wins vs. 8 losses) and IMTCMO (10 wins vs. 6 losses).

The mechanism-level diagnosis reveals a fundamental distinction between artificial benchmarks and real-world problems. Artificial benchmarks typically construct difficulty using deceptive mathematical functions. In contrast, the difficulty in RWMOPs stems from multi-physics equations with vastly heterogeneous magnitudes and units. For instance, a mechanical design problem might simultaneously restrict a physical deflection to less than $0.05$ mm while bounding an elasticity modulus limit at $30 \times 10^6$. When traditional algorithms aggregate these constraint violations into a single $CV(X)$, the constraints with massive numerical magnitudes completely overshadow the subtle, tighter constraints, effectively blinding the search direction. 

RCCMO's fundamental architecture entirely circumvents this scale-imbalance issue. By explicitly isolating constraints into dedicated dual populations during Stage 2, RCCMO evaluates and targets the violation of one specific constraint at a time. This structural isolation ensures that numerically small but geometrically critical constraints are strictly respected rather than being washed out by an aggregated penalty function. The effectiveness of this mechanism is vividly demonstrated by RCCMO's consistent superiority across the RWMOP suite, where it reliably secures higher relative HV values across highly heterogeneous problem instances. By safely navigating the extremely narrow and heavily restricted feasible tubes typical in real-world multi-physics optimizations, RCCMO yields high-quality, well-distributed feasible solutions where generic aggregated CMOEAs consistently fail.

\begin{table*}[htbp]
  \centering
  \caption{Statistical Summary of HV Values ($+$ / $-$ / $\approx$) and Average Ranks across RWMOPs.}
  \label{tab:summary_HV}
  \resizebox{\textwidth}{!}{
  \begin{tabular}{lcccccccc}
    \toprule
    Problem Suite & APSEA & C3M & FCDSU & IMTCMO & MCCMO & MSCMO & MTOTC & RCCMO \\
    \midrule
    RWMOP ($+$/$-$/$\approx$) & 5/15/8 & 5/15/8 & 5/17/6 & 6/10/12 & 7/15/6 & 2/20/6 & 8/13/7 & - \\
    \midrule
    \quad\textit{Avg. Rank} & 4.61 & 4.46 & 5.32 & 4.18 & 4.43 & 5.93 & 3.86 & \textbf{3.21} \\
    \bottomrule
    \multicolumn{9}{l}{\textit{Note:} Nemenyi test Critical Difference (CD) = 1.98 at $\alpha=0.05$.}\\
  \end{tabular}
  }
\end{table*}

\begin{table*}[htbp]
  \centering
  \caption{Average Running Time (in seconds) and Standard Deviation across Test Suites. The shortest time is \textbf{bolded}, and the second shortest is \underline{underlined}.}
  \label{tab:runtime_summary}
  \resizebox{\textwidth}{!}{
  \begin{tabular}{lccccccccc}
    \toprule
    Problem Suite & APSEA & C3M & FCDS & IMTCMO & MCCMO & MSCMO & MTOTC & RCCMO w/o AUS & RCCMO \\
    \midrule
    DASCMOP & \textbf{7.74 (1.32)} & 29.12 (7.73) & 43.68 (2.99) & 14.66 (6.63) & 45.20 (29.62) & \underline{13.14 (5.33)} & 64.59 (18.79) & 22.05 (5.67) & 16.62 (5.33) \\
    DOC & \textbf{14.04 (3.94)} & 35.69 (30.37) & 46.02 (16.95) & \underline{17.07 (5.43)} & 25.15 (10.98) & 33.04 (10.47) & 70.96 (36.81) & 25.51 (4.07) & 25.60 (4.84) \\
    LIRCMOP & \textbf{9.54 (4.10)} & 13.25 (4.47) & 61.60 (18.83) & 21.84 (11.42) & 22.02 (11.12) & \underline{12.22 (3.79)} & 45.09 (35.91) & 19.62 (2.23) & 19.15 (2.22) \\
    SDC & \textbf{11.76 (3.14)} & 13.75 (4.32) & 39.00 (12.30) & 13.66 (3.64) & 20.93 (5.63) & \underline{13.57 (11.77)} & 29.21 (16.41) & 20.02 (2.88) & 19.66 (3.56) \\
    ZXH\_CF & \underline{14.98 (4.85)} & 19.46 (6.83) & 73.25 (13.64) & 24.69 (6.19) & 33.76 (13.95) & \textbf{14.36 (4.10)} & 92.00 (30.72) & 24.31 (5.58) & 23.31 (6.44) \\
    \midrule
    \textbf{Overall Average} & \textbf{11.61} & 22.25 & 52.71 & 18.38 & 29.41 & \underline{17.26} & 60.37 & 22.30 & 20.87 \\
    \bottomrule
  \end{tabular}
  }
\end{table*}

\subsection{Complexity and Runtime Analysis}
\label{1f1}

In this section, we analyze the theoretical computational complexity of the core constraint-handling mechanisms proposed in RCCMO and evaluate its empirical running time against the compared algorithms.

Suppose $N$ is the population size, $M$ is the number of objectives, $Nc$ is the number of constraints, and $V$ is the update interval for inactive constraint populations. The computational overhead of RCCMO primarily stems from three newly introduced components: the probe population update, the dynamic priority determination, and the environmental selection of the dual-auxiliary populations.

First, for the probe population ($Pb$), the primary computational overhead involves non-dominated sorting and crowding distance calculation based on negative objectives and modified constraint violations, which results in a worst-case complexity of $O(M N^2)$ per generation. 

Second, the dynamic priority determination (Algorithm \ref{A4}) requires calculating the feasibility and blockage rates across the populations, taking $O(Nc \cdot N)$ operations, followed by a fast sorting operation that takes $O(Nc \log Nc)$. Since $Nc \ll N$, this ranking overhead is mathematically negligible.

Finally, the most computationally intensive operation in any multi-population framework is environmental selection. If a naive parallel approach were employed to update all $2Nc+2$ populations in every single generation, it would incur an unacceptable, geometrically scaling overhead of $O(Nc \cdot M N^2)$. Crucially, RCCMO mitigates this via the Asymmetric Update Strategy (AUS). Under AUS, only the main population, the probe population, and the currently active constraint populations (at most two) are updated generation-by-generation. The remaining inactive constraint populations are frozen and updated only once every $V$ generations. This mechanism gracefully reduces the amortized non-evaluative complexity for the auxiliary populations to $O(\frac{Nc}{V} \cdot M N^2)$. Therefore, despite maintaining a comprehensive geometric map of all constraints, the theoretical complexity of RCCMO remains strictly bounded and scales highly efficiently with the number of constraints.

To empirically validate this theoretical efficiency, Table \ref{tab:runtime_summary} presents the average running time and standard deviation of all algorithms. As observed, single-population methods without complex archiving or parallel evaluation mechanisms, such as APSEA (11.61s) and MSCMO (17.26s), naturally run the fastest. In stark contrast, methods relying on maintaining multiple parallel populations or heavy constraint decomposition, including MTOTC (60.37s), FCDS (52.71s), and MCCMO (29.41s), exhibit massive non-evaluative computational overhead. 

Impressively, despite managing $2Nc+2$ populations to meticulously map the geometric topology, RCCMO achieves an overall average running time of 20.87s, remaining highly competitive and significantly outpacing heavier multi-population frameworks. Furthermore, the ablation comparison directly verifies the necessity of the AUS mechanism. The variant without AUS (\textit{w/o AUS}) records a higher average runtime of 22.30s. The introduction of AUS explicitly trims redundant sorting operations, demonstrating that RCCMO's structural complexity scales efficiently without bogging down the execution speed.

\begin{figure}[!htbp]
	\setlength{\abovecaptionskip}{0pt}
	\setlength{\belowcaptionskip}{0pt}
	\centering
	\includegraphics[width=\textwidth]{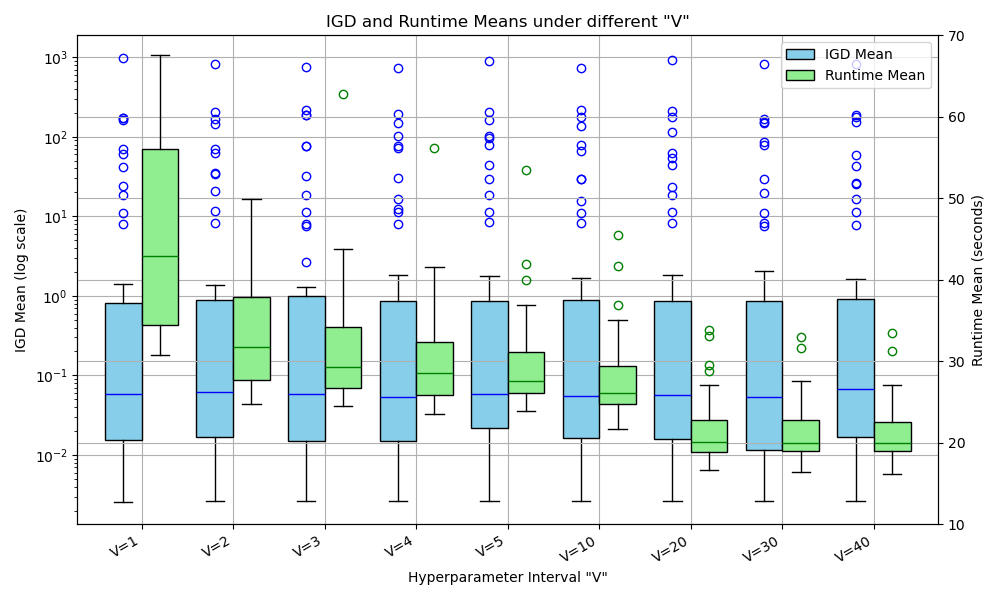}
	\caption{Comparison of the performance and running time of RCCMO under different interval parameter $V$.}
	\label{para}
\end{figure}

To further investigate the impact of the interval parameter $V$, we analyzed the running time and performance (IGD distribution) of RCCMO under different $V$ settings, as illustrated in Fig. \ref{para}. The parameter $V$ controls the update frequency of the non-active constraint populations, which indirectly affects the timeliness of the topological data used for priority judgment. As $V$ increases, the algorithm's running time significantly decreases due to fewer environmental selection executions. Interestingly, the performance bounds remain highly stable up to $V=30$, demonstrating that updating topological information every 30 generations is sufficient to ensure accurate constraint prioritization. However, beyond $V > 20$, the marginal improvement in execution speed diminishes. Therefore, $V=30$ is selected as the optimal default setting to strike a perfect balance between robust geometric mapping and optimal computational efficiency.

\begin{table*}[htbp]
  \centering
  \caption{Statistical Summary of IGD Values ($+$ / $-$ / $\approx$) and Average Ranks for Ablation Variants across Test Suites.}
  \label{tab:ablation}
  \resizebox{\textwidth}{!}{
  \begin{tabular}{lccccccc}
    \toprule
    Problem Suite & w/o AUS & Only-Neg & w/o Re-rank & w/o Flip & Only-Pos & Rand-Priority & RCCMO (Baseline) \\
    \midrule
    DASCMOP ($+$/$-$/$\approx$) & 0/0/9 & 0/0/9 & 1/2/6 & 0/0/9 & 0/2/7 & 1/3/5 & - \\
    \quad\textit{Avg. Rank} & \textbf{3.44} & 3.67 & 4.56 & 4.56 & 3.44 & 4.22 & 4.11 \\
    \midrule
    DOC ($+$/$-$/$\approx$) & 1/0/8 & 1/1/7 & 0/2/7 & 0/1/8 & 0/6/3 & 0/2/7 & - \\
    \quad\textit{Avg. Rank} & 3.33 & 3.67 & 3.56 & 4.00 & 6.33 & 4.33 & \textbf{2.78} \\
    \midrule
    LIRCMOP ($+$/$-$/$\approx$) & 0/1/13 & 0/8/6 & 0/1/13 & 1/1/12 & 2/4/8 & 1/2/11 & - \\
    \quad\textit{Avg. Rank} & 4.07 & 5.50 & 3.64 & 4.14 & 3.86 & 3.50 & \textbf{3.29} \\
    \midrule
    SDC ($+$/$-$/$\approx$) & 1/1/13 & 0/3/12 & 0/1/14 & 1/1/13 & 3/9/3 & 1/1/13 & - \\
    \quad\textit{Avg. Rank} & 3.73 & 3.87 & 4.07 & 3.67 & 5.67 & 3.80 & \textbf{3.20} \\
    \midrule
    ZXH\_CF ($+$/$-$/$\approx$) & 1/0/15 & 0/9/7 & 0/3/13 & 2/1/13 & 1/1/14 & 1/4/11 & - \\
    \quad\textit{Avg. Rank} & \textbf{3.19} & 5.56 & 4.31 & 3.88 & 3.38 & 4.00 & 3.69 \\
    \midrule
    \textbf{Overall} ($+$/$-$/$\approx$) & \textbf{3/2/58} & \textbf{1/21/41} & \textbf{1/9/53} & \textbf{4/4/55} & \textbf{6/22/35} & \textbf{4/12/47} & \textbf{Baseline} \\
    \textbf{Overall Avg. Rank} & \textbf{3.57} & \textbf{4.60} & \textbf{4.03} & \textbf{4.00} & \textbf{4.46} & \textbf{3.92} & \textbf{3.41} \\
    \bottomrule
    \multicolumn{8}{l}{\textit{Note:} Nemenyi test Critical Difference (CD) = 1.13 at $\alpha=0.05$.}\\
  \end{tabular}
  }
\end{table*}

\subsection{Ablation Study}
\label{sec:ablation}

To rigorously validate the necessity and individual contributions of the core mechanisms proposed in RCCMO, we designed six ablation variants. The statistical comparison between these variants and the full RCCMO baseline across the 63 test instances is summarized in Table \ref{tab:ablation}.

\textbf{1. Validation of Dual-Directional Search (\textit{Only-Pos} and \textit{Only-Neg}):} 
The \textit{Only-Pos} variant restricts all constraint handling exclusively to the evolutionary direction (locating SCPFs), completely disabling the anti-evolutionary boundary mapping. Conversely, the \textit{Only-Neg} variant forces all active constraints to be searched from the anti-evolutionary direction to map infeasible boundaries. As evidenced by Table \ref{tab:ablation}, both variants suffer catastrophic performance degradation, losing to the baseline on 22 and 21 instances, respectively. \textit{Only-Pos} specifically collapses on constraint-dense suites like DOC and SDC, proving that merely searching for isolated CPFs fails when the feasible region is severely blocked. This confirms that assigning distinct search directions based on the topological role of each constraint is fundamentally required.

\textbf{2. Validation of Dynamic Prioritization (\textit{Rand-Priority} and \textit{w/o Re-rank}):} 
The \textit{Rand-Priority} variant replaces the statistical sorting mechanism with completely random constraint prioritization. Its severe performance drop (12 losses, Avg Rank: 3.92) confirms the necessity of our geometry-based ranking framework. However, even with initial sorting, early topological inferences are susceptible to stochastic noise. The \textit{w/o Re-rank} variant calculates the priority only once at the end of Stage 1 and locks it permanently. This static approach leads to 9 losses against the baseline (Avg Rank: 4.03), validating that cyclical, dynamic re-evaluation is crucial to self-correct early misjudgments as the populations converge.

\textbf{3. Validation of Real-Time Correction (\textit{w/o Flip}):} 
The \textit{w/o Flip} variant disables the Instant Bi-directional Flipping mechanism, forcing the algorithm to wait until the next formal stage transition to change a constraint's search direction. While its overall average rank (4.00) is slightly better than the static variants, it still loses on 4 critical instances. This proves that real-time geometric responsiveness—instantly flipping to exploit an unexpectedly discovered feasible pocket, or immediately retreating to map an impenetrable wall—provides a vital topological safeguard in highly deceptive landscapes.

\textbf{4. Validation of Asymmetric Update Strategy (\textit{w/o AUS}):} 
The \textit{w/o AUS} variant forces all dual auxiliary populations ($P_i^{pos}$ and $P_i^{neg}$) to execute environmental selection in every single generation. Statistically, its optimization performance is highly similar to the baseline (58 ties), yielding a very close average rank (3.57 vs. 3.41). This is completely aligned with our theoretical design: AUS is primarily a computational efficiency mechanism, not an optimization booster. While freezing inactive populations slightly alters the diversity maintenance (causing minor fluctuations in 5 instances), the full RCCMO baseline maintains superior optimization capability while saving valuable computational overhead (as demonstrated in Table \ref{tab:runtime_summary}). Thus, AUS serves as an ideal bridge between topological mapping accuracy and execution speed.

\section{Conclusion and Future Work}
\label{sec:conclusion}

In this study, we fundamentally reconsidered the role of constraints in multi-objective optimization. Rather than treating constraint satisfaction as a monolithic, aggregated mathematical penalty, RCCMO shifts the paradigm by approaching it as a geometric topology problem. By explicitly distinguishing between constraints that directly shape the final Constrained Pareto Front (CPF) and those that merely act as obstructive infeasible boundaries, we established a novel framework based on constraint isolation and dynamic prioritization. This topological perspective is particularly valuable for real-world engineering applications, where physical constraints often possess vastly heterogeneous scales and units. By completely decoupling these constraints, RCCMO successfully immunizes the evolutionary search against deceptive, scale-imbalanced landscapes, providing a highly transparent and robust methodology for navigating severely compressed feasible regions.

The successful realization of this geometric insight heavily relies on the synergy of our proposed architectural mechanisms. By assigning a dedicated dual-population structure to each constraint, RCCMO is capable of executing targeted searches from both evolutionary and anti-evolutionary directions. Moreover, recognizing that early statistical inferences are inherently heuristic, the algorithm is fortified with an Instant Bi-directional Flipping mechanism, enabling it to act as a vigilant geometric monitor and self-correct directional misjudgments in real time. Crucially, we demonstrated that maintaining a comprehensive topological map does not inevitably lead to computational paralysis. The introduction of the Asymmetric Update Strategy (AUS) elegantly circumvents the severe non-evaluative computational bottlenecks that have long plagued cooperative multi-population frameworks. Supported by rigorous statistical analyses across 63 benchmarks and 29 complex real-world problems, RCCMO proves that exceptional topological mapping accuracy can be achieved synchronously with state-of-the-art optimization performance and outstanding execution efficiency.

Despite its highly competitive performance, explicitly tracing constraint boundaries through dual-directional search naturally consumes function evaluations, opening several promising avenues for future research. For computationally expensive real-world engineering tasks, integrating lightweight surrogate models or machine learning classifiers to approximate these active infeasible boundaries could significantly reduce the evaluation burden while maintaining geometric precision. Furthermore, as the current probe mechanism relies on strict Pareto dominance to detect obstructive boundaries, extending RCCMO to Constrained Many-Objective Optimization Problems (CMaOPs) will require adapting this detection strategy with reference vectors or indicator-based criteria to prevent the loss of selection pressure in high-dimensional objective spaces. Finally, exploring how the decoupled dual-population architecture can be tailored to swiftly detect moving boundaries and re-anchor search trajectories in dynamic or uncertain environments presents a highly practical direction for advancing evolutionary constraint-handling techniques.

\bibliography{sample}

@article{apsea,
  title={Adaptive population sizing for multi-population based constrained multi-objective optimization},
  author={Tian, Ye and Wang, Ruiqin and Zhang, Yajie and Zhang, Xingyi},
  journal={Neurocomputing},
  volume={621},
  pages={129296},
  year={2025},
  publisher={Elsevier}
}

@article{chu2024competitive,
  title={Competitive multitasking for computational resource allocation in evolutionary constrained multi-objective optimization},
  author={Chu, Xiaoliang and Ming, Fei and Gong, Wenyin},
  journal={IEEE Transactions on Evolutionary Computation},
  year={2024},
  publisher={IEEE}
}

@article{liu2024coevolutionary,
  title={A coevolutionary algorithm with detection and supervision strategies for constrained multiobjective optimization},
  author={Liu, Shaoning and Feng, Jian and Yang, Shengxiang and Zheng, Jun and Xiao, Qi},
  journal={IEEE Transactions on Evolutionary Computation},
  year={2024},
  publisher={IEEE}
}

@article{DPCPRA,
  title={A dual-population evolutionary algorithm based on dynamic constraint processing and resources allocation for constrained multi-objective optimization problems},
  author={Qiao, Kangjia and Chen, Zhaolin and Qu, Boyang and Yu, Kunjie and Yue, Caitong and Chen, Ke and Liang, Jing},
  journal={Expert Systems with Applications},
  volume={238},
  pages={121707},
  year={2024},
  publisher={Elsevier}
}

@article{liu2025constraint,
  title={Constraint-Pareto Dominance and Diversity Enhancement Strategy based Evolutionary Algorithm for Solving Constrained Multiobjective Optimization Problems},
  author={Liu, Zhe and Han, Fei and Ling, Qinghua and Han, Henry and Jiang, Jing},
  journal={IEEE Transactions on Evolutionary Computation},
  year={2025},
  publisher={IEEE}
}

@article{mg,
  title={Computer-aided multi-objective optimization in small molecule discovery},
  author={Fromer, Jenna C and Coley, Connor W},
  journal={Patterns},
  volume={4},
  number={2},
  year={2023},
  publisher={Elsevier}
}

@article{schd,
  title={A multi-objective optimization for resource allocation of emergent demands in cloud computing},
  author={Chen, Jing and Du, Tiantian and Xiao, Gongyi},
  journal={Journal of Cloud Computing},
  volume={10},
  pages={1--17},
  year={2021},
  publisher={Springer}
}

@article{imtcmo,
  title={Evolutionary constrained multiobjective optimization: Scalable high-dimensional constraint benchmarks and algorithm},
  author={Qiao, Kangjia and Liang, Jing and Yu, Kunjie and Yue, Caitong and Lin, Hongyu and Zhang, Dezheng and Qu, Boyang},
  journal={IEEE Transactions on Evolutionary Computation},
  year={2023},
  publisher={IEEE}
}

@article{mtotc,
  title={Decoupling Constraint: Task Clone-Based Multi-Tasking Optimization for Constrained Multi-Objective Optimization},
  author={Li, Genghui and Wang, Zhenkun and Gao, Weifeng and Wang, Ling},
  journal={IEEE Transactions on Evolutionary Computation},
  year={2024},
  publisher={IEEE}
}

@article{c3m,
  title={A multi-stage algorithm for solving multi-objective optimization problems with multi-constraints},
  author={Sun, Ruiqing and Zou, Juan and Liu, Yuan and Yang, Shengxiang and Zheng, Jinhua},
  journal={IEEE Transactions on Evolutionary Computation},
  year={2022},
  publisher={IEEE}
}

@article{cmoes,
  title={Two-stage multi-objective evolution strategy for constrained multi-objective optimization},
  author={Zhang, Kai and Xu, Zhiwei and Yen, Gary G and Zhang, Ling},
  journal={IEEE Transactions on Evolutionary Computation},
  year={2022},
  publisher={IEEE}
}

@article{cmoeapp,
  title={Cooperative multiobjective evolutionary algorithm with propulsive population for constrained multiobjective optimization},
  author={Wang, Jiahai and Li, Yanyue and Zhang, Qingfu and Zhang, Zizhen and Gao, Shangce},
  journal={IEEE Transactions on Systems, Man, and Cybernetics: Systems},
  volume={52},
  number={6},
  pages={3476--3491},
  year={2021},
  publisher={IEEE}
}

@article{cmoqlmt,
  title={Adaptive auxiliary task selection for multitasking-assisted constrained multi-objective optimization [feature]},
  author={Ming, Fei and Gong, Wenyin and Gao, Liang},
  journal={IEEE Computational Intelligence Magazine},
  volume={18},
  number={2},
  pages={18--30},
  year={2023},
  publisher={IEEE}
}

@article{mtcmo,
  title={Dynamic auxiliary task-based evolutionary multitasking for constrained multi-objective optimization},
  author={Qiao, Kangjia and Yu, Kunjie and Qu, Boyang and Liang, Jing and Song, Hui and Yue, Caitong and Lin, Hongyu and Tan, Kay Chen},
  journal={IEEE Transactions on Evolutionary Computation},
  year={2022},
  publisher={IEEE}
}

@article{cmoemt,
  title={Constrained multi-objective optimization via multitasking and knowledge transfer},
  author={Ming, Fei and Gong, Wenyin and Wang, Ling and Gao, Liang},
  journal={IEEE Transactions on Evolutionary Computation},
  year={2022},
  publisher={IEEE}
}

@article{dbemto,
  title={A self-adaptive evolutionary multi-task based constrained multi-objective evolutionary algorithm},
  author={Qiao, Kangjia and Liang, Jing and Yu, Kunjie and Wang, Minghui and Qu, Boyang and Yue, Caitong and Guo, Yinan},
  journal={IEEE Transactions on Emerging Topics in Computational Intelligence},
  year={2023},
  publisher={IEEE}
}

@article{dbccmoea,
  title={A dual-population based bidirectional coevolution algorithm for constrained multi-objective optimization problems},
  author={Bao, Qian and Wang, Maocai and Dai, Guangming and Chen, Xiaoyu and Song, Zhiming and Li, Shuijia},
  journal={Expert Systems with Applications},
  volume={215},
  pages={119258},
  year={2023},
  publisher={Elsevier}
}

@article{cmaoo,
  title={A constrained multi-objective evolutionary algorithm assisted by an additional objective function},
  author={Yang, Yongkuan and Huang, Pei-Qiu and Kong, Xiangsong and Zhao, Jing},
  journal={Applied Soft Computing},
  volume={132},
  pages={109904},
  year={2023},
  publisher={Elsevier}
}

@article{cmoeatcp,
  title={A constrained multi-objective optimization algorithm with two cooperative populations},
  author={Zhang, Jianlin and Cao, Jie and Zhao, Fuqing and Chen, Zuohan},
  journal={Memetic Computing},
  volume={14},
  number={1},
  pages={95--113},
  year={2022},
  publisher={Springer}
}

@article{tscso,
  title={A tri-stage competitive swarm optimizer for constrained multi-objective optimization},
  author={Dong, Jun and Gong, Wenyin and Ming, Fei},
  journal={Applied Intelligence},
  volume={53},
  number={7},
  pages={7892--7916},
  year={2023},
  publisher={Springer}
}

@article{zx1,
  title={Multi-objective evolutionary computation for topology coverage assessment problem},
  author={Zhou, Xing and Wang, Huaimin and Ding, Bo and Peng, Wei and Wang, Rui},
  journal={Knowledge-Based Systems},
  volume={177},
  pages={1--10},
  year={2019},
  publisher={Elsevier}
}

@article{zx2,
  title={Solving multi-scenario cardinality constrained optimization problems via multi-objective evolutionary algorithms},
  author={Zhou, Xing and Wang, Huaimin and Peng, Wei and Ding, Bo and Wang, Rui},
  journal={Science China Information Sciences},
  volume={62},
  pages={1--18},
  year={2019},
  publisher={Springer}
}

@ARTICLE{bico,
	author={Liu, Zhi-Zhong and Wang, Bing-Chuan and Tang, Ke},
	journal={IEEE Transactions on Cybernetics}, 
	title={Handling Constrained Multiobjective Optimization Problems via Bidirectional Coevolution}, 
	year={2022},
	volume={52},
	number={10},
	pages={10163-10176},
	doi={10.1109/TCYB.2021.3056176}}

@article{nsga2,
  title={A fast and elitist multiobjective genetic algorithm: NSGA-II},
  author={Deb, Kalyanmoy and Pratap, Amrit and Agarwal, Sameer and Meyarivan, TAMT},
  journal={IEEE Transactions on Evolutionary Computation},
  volume={6},
  number={2},
  pages={182--197},
  year={2002},
  publisher={IEEE}
}

@inproceedings{ep1,
  title={Constrained optimization by the $\epsilon$ constrained differential evolution with an archive and gradient-based mutation},
  author={ Takahama, Tetsuyuki  and  Sakai, Setsuko },
  booktitle={IEEE Congress on Evolutionary Computation},
  pages={1-9},
  year={2010},
}

@article{MSCMO,
	author = {Ma, Haiping and Wei, Haoyu and Tian, Ye and Cheng, Ran and Zhang, Xingyi},
	year = {2021},
	month = {01},
	pages = {},
	title = {A multi-stage evolutionary algorithm for multi-objective optimization with complex constraints},
	volume = {560},
	journal = {Information Sciences},
	doi = {10.1016/j.ins.2021.01.029}
}

@ARTICLE{igdm,  author={Bosman, P.A.N. and Thierens, D.},  journal={IEEE Transactions on Evolutionary Computation},   title={The balance between proximity and diversity in multiobjective evolutionary algorithms},   year={2003},  volume={7},  number={2},  pages={174-188},  doi={10.1109/TEVC.2003.810761}}

@article{HV,
	author = {While, Lyndon and Hingston, Philip and Barone, Luigi and Huband, Simon},
	year = {2006},
	month = {03},
	pages = {29 - 38},
	title = {A faster algorithm for calculating hypervolume},
	volume = {10},
	journal = {IEEE Transactions on Evolutionary Computation},
	doi = {10.1109/TEVC.2005.851275}
}

@article{cmosma,
  title={A self-organizing map approach for constrained multi-objective optimization problems},
  author={He, Chao and Li, Ming and Zhang, Congxuan and Chen, Hao and Zhong, Peilong and Li, Zhengxiu and Li, Junhua},
  journal={Complex \& Intelligent Systems},
  volume={8},
  number={6},
  pages={5355--5375},
  year={2022},
  publisher={Springer}
}

@article{mscea,
  title={Design and analysis of helper-problem-assisted evolutionary algorithm for constrained multiobjective optimization},
  author={Zhang, Yajie and Tian, Ye and Jiang, Hao and Zhang, Xingyi and Jin, Yaochu},
  journal={Information Sciences},
  volume={648},
  pages={119547},
  year={2023},
  publisher={Elsevier}
}

@inproceedings{spf1,
  title={A self adaptive penalty function based algorithm for constrained optimization},
  author={Tessema, Biruk and Yen, Gary G},
  booktitle={2006 IEEE international conference on evolutionary computation},
  pages={246--253},
  year={2006},
  organization={IEEE}
}

@article{ddcmoea,
  title={A Novel Dual-Stage Dual-Population Evolutionary Algorithm for Constrained Multiobjective Optimization},
  author={Ming, Mengjun and Wang, Rui and Ishibuchi, Hisao and Zhang, Tao},
  journal={IEEE Transactions on Evolutionary Computation},
  volume={26},
  number={5},
  pages={1129--1143},
  year={2021},
  publisher={IEEE}
}

@article{llh1,
  title={Accelerating large-scale multiobjective optimization via problem reformulation},
  author={He, Cheng and Li, Lianghao and Tian, Ye and Zhang, Xingyi and Cheng, Ran and Jin, Yaochu and Yao, Xin},
  journal={IEEE Transactions on Evolutionary Computation},
  volume={23},
  number={6},
  pages={949--961},
  year={2019},
  publisher={IEEE}
}

@article{zxhcf,
  title={Constrained multiobjective optimization: Test problem construction and performance evaluations},
  author={Zhou, Yuren and Xiang, Yi and He, Xiaoyu},
  journal={IEEE Transactions on Evolutionary Computation},
  volume={25},
  number={1},
  pages={172--186},
  year={2020},
  publisher={IEEE}
}

@article{mccmo,
  title={A Multi-Population Evolutionary Algorithm Using New Cooperative Mechanism for Solving Multi-Objective Problems with Multi-Constraint},
  author={Zou, Juan and Sun, Ruiqing and Liu, Yuan and Hu, Yaru and Yang, Shengxiang and Zheng, Jinhua and Li, Ke},
  journal={IEEE Transactions on Evolutionary Computation},
  year={2023},
  publisher={IEEE}
}

@article{trip,
  title={A tri-population based co-evolutionary framework for constrained multi-objective optimization problems},
  author={Ming, Fei and Gong, Wenyin and Wang, Ling and Lu, Chao},
  journal={Swarm and Evolutionary Computation},
  volume={70},
  pages={101055},
  year={2022},
  publisher={Elsevier}
}

@article{mob,
  title={A multiobjective optimization-based evolutionary algorithm for constrained optimization},
  author={Cai, Zixing and Wang, Yong},
  journal={IEEE Transactions on evolutionary computation},
  volume={10},
  number={6},
  pages={658--675},
  year={2006},
  publisher={IEEE}
}

@article{sr,
  title={Stochastic ranking for constrained evolutionary optimization},
  author={Runarsson, Thomas P. and Yao, Xin},
  journal={IEEE Transactions on evolutionary computation},
  volume={4},
  number={3},
  pages={284--294},
  year={2000},
  publisher={IEEE}
}

@article{tsti,
  title={A two-stage evolutionary algorithm based on three indicators for constrained multi-objective optimization},
  author={Dong, Jun and Gong, Wenyin and Ming, Fei and Wang, Ling},
  journal={Expert Systems with Applications},
  volume={195},
  pages={116499},
  year={2022},
  publisher={Elsevier}
}

@article{2017PlatEMO,
	author = {Tian, Ye and Cheng, Ran and Zhang, Xingyi and Jin, Yaochu},
	year = {2017},
	month = {11},
	pages = {73-87},
	title = {PlatEMO: A MATLAB Platform for Evolutionary Multi-Objective Optimization},
	volume = {12},
	journal = {IEEE Computational Intelligence Magazine},
	doi = {10.1109/MCI.2017.2742868}
}

@article{pps,
	author = {Fan, Zhun and Li, Wenji and Cai, Xinye and Li, Hui and Wei, Caimin and Zhang, Qingfu and Deb, Kalyanmoy and Goodman, Erik},
	year = {2017},
	month = {09},
	pages = {},
	title = {Push and Pull Search for Solving Constrained Multi-objective Optimization Problems},
	volume = {44},
	journal = {Swarm and Evolutionary Computation},
	doi = {10.1016/j.swevo.2018.08.017}
}

@article{PIC_2025,
  title={Constrained Multiobjective Evolutionary Optimization With Population Image Convolution},
  author={Wu, Wenhao and others},
  journal={IEEE Transactions on Systems, Man, and Cybernetics: Systems},
  volume={55},
  number={11},
  pages={7826--7840},
  year={2025},
  publisher={IEEE}
}

@article{SSMOEA_2025,
  title={A Subspace Search-Based Evolutionary Algorithm for Large-Scale Constrained Multiobjective Optimization and Application},
  author={Ban, Xuanxuan and Liang, Jing and Yu, Kunjie and Wang, Yaonan},
  journal={IEEE Transactions on Cybernetics},
  volume={55},
  number={5},
  pages={2486--2499},
  year={2025},
  publisher={IEEE}
}

@article{CMOEATA_2025,
  title={Two-Stage Archive Evolutionary Algorithm for Constrained Multi-Objective Optimization},
  author={Zhang, Kai and Zhao, Siyuan and Zeng, Hui and Chen, Junming},
  journal={Mathematics},
  volume={13},
  number={3},
  pages={470},
  year={2025},
  publisher={MDPI}
}

@article{CPTSEA_2025,
  title={Competition-based two-stage evolutionary algorithm for constrained multi-objective optimization},
  author={Hao, Lupeng and Peng, Weihang and Liu, Junhua and Zhang, Wei and Li, Yuan and Qin, Kaixuan},
  journal={Mathematics and Computers in Simulation},
  volume={230},
  pages={207--226},
  year={2025},
  publisher={Elsevier}
}

@article{MTMOEA_2025,
  title={A multi-task multi-objective evolutionary algorithm for constrained multi-objective optimization problems},
  author={Liu, Tianyu and others},
  journal={Memetic Computing},
  volume={17},
  number={3},
  year={2025},
  publisher={Springer}
}

@article{CIDEMT_2026,
  title={Constraint Intensity-Driven Evolutionary Multitasking for Constrained Multi-Objective Optimization},
  author={Zheng, Leyu and Xiao, Mingming and Ren, Yi and Li, Ke and Sun, Chang},
  journal={Computers, Materials \& Continua},
  year={2026}
}

@article{CCMT_2025,
  title={A collaborative competition multitasking framework for constrained multi-objective optimization},
  author={Feng, Xinyu and others},
  journal={Expert Systems with Applications}, 
  year={2025}
}

@article{cmoeams,
	author = {Tian, Ye and Zhang, Yajie and Su, Yansen and Zhang, Xingyi and Tan, Kay and Jin, Yaochu},
	year = {2020},
	month = {08},
	pages = {},
	title = {Balancing Objective Optimization and Constraint Satisfaction in Constrained Evolutionary Multi-Objective Optimization},
	journal = {IEEE Transactions on Cybernetics},
	doi = {10.1109/TCYB.2020.3021138}
}

@ARTICLE{cdpea,  author={Ming, Mengjun and Trivedi, Anupam and Wang, Rui and Srinivasan, Dipti and Zhang, Tao},  journal={IEEE Transactions on Evolutionary Computation},   title={A Dual-Population-Based Evolutionary Algorithm for Constrained Multiobjective Optimization},   year={2021},  volume={25},  number={4},  pages={739-753},  doi={10.1109/TEVC.2021.3066301}}

@article{top,
  title={Handling constrained multiobjective optimization problems with constraints in both the decision and objective spaces},
  author={Liu, Zhi-Zhong and Wang, Yong},
  journal={IEEE Transactions on Evolutionary Computation},
  volume={23},
  number={5},
  pages={870--884},
  year={2019},
  publisher={IEEE}
}

@article{fcds,
  title   = {Fuzzy Constraint Dominance Strategy for Constrained Multiobjective Optimization Problems With Multiple Constraints},
  journal = {IEEE/CAA Journal of Automatica Sinica},
  volume  = {12},
  number  = {1},
  pages   = {1--14},
  year    = {2025},
  issn    = {2329-9266},
  doi     = {10.1109/JAS.2025.125255},
  author  = {Weixiong Huang and Rui Wang and Tao Zhang and Sheng Qi and Ling Wang}
}

@article{lir,
	author = {Fan, Zhun and Li, Wenji and Cai, Xinye and Huang, Han and Fang, Yi and Yugen, You and Mo, Jiajie and Wei, Caimin and Goodman, Erik},
	year = {2019},
	month = {12},
	pages = {},
	title = {An Improved Epsilon Constraint-handling Method in MOEA/D for CMOPs with Large Infeasible Regions},
	volume = {23},
	journal = {Soft Computing},
	doi = {10.1007/s00500-019-03794-x}
}

@article{dascmop,
	author = {Fan, Zhun and Li, Wenji and Cai, Xinye and Li, Hui and Zhang, Qingfu and Deb, Kalyan and Goodman, Erik},
	year = {2019},
	month = {05},
	pages = {1-28},
	title = {Difficulty Adjustable and Scalable Constrained Multi-objective Test Problem Toolkit},
	volume = {28},
	journal = {Evolutionary Computation},
	doi = {10.1162/evco_a_00259}
}

@article{rwmop,
  title={A benchmark-suite of real-world constrained multi-objective optimization problems and some baseline results},
  author={Kumar, Abhishek and Wu, Guohua and Ali, Mostafa Z and Luo, Qizhang and Mallipeddi, Rammohan and Suganthan, Ponnuthurai Nagaratnam and Das, Swagatam},
  journal={Swarm and Evolutionary Computation},
  volume={67},
  pages={100961},
  year={2021},
  publisher={Elsevier}
}

@article{CCMO,
  title={A coevolutionary framework for constrained multiobjective optimization problems},
  author={Tian, Ye and Zhang, Tao and Xiao, Jianhua and Zhang, Xingyi and Jin, Yaochu},
  journal={IEEE Transactions on Evolutionary Computation},
  volume={25},
  number={1},
  pages={102--116},
  year={2020},
  publisher={IEEE}
}

@ARTICLE{emcmo,  author={Qiao, Kangjia and Yu, Kunjie and Qu, Boyang and Liang, Jing and Song, Hui and Yue, Caitong},  journal={IEEE Transactions on Evolutionary Computation},   title={An Evolutionary Multitasking Optimization Framework for Constrained Multiobjective Optimization Problems},   year={2022},  volume={26},  number={2},  pages={263-277},  doi={10.1109/TEVC.2022.3145582}}

@ARTICLE{ctaea,  author={Li, Ke and Chen, Renzhi and Fu, Guangtao and Yao, Xin},  journal={IEEE Transactions on Evolutionary Computation},   title={Two-Archive Evolutionary Algorithm for Constrained Multiobjective Optimization},   year={2019},  volume={23},  number={2},  pages={303-315},  doi={10.1109/TEVC.2018.2855411}}

\end{document}